\documentclass[preprint,12pt]{elsarticle}
\usepackage{epsfig}
\usepackage{amssymb}
\usepackage{amsmath}
\usepackage{amsthm}
\usepackage{bm}
\usepackage{caption}
\usepackage{subcaption}
\usepackage{mathabx}
\newcommand{\ignore}[1]{}
\usepackage{lipsum}
\makeatletter
\def\ps@pprintTitle{%
 \let\@oddhead\@empty
 \let\@evenhead\@empty
 \def\@oddfoot{}%
 \let \@oddfoot}
\makeatother
\begin{document}

\begin{frontmatter}

\title{Interpretable Convolutional Neural Networks via Feedforward Design}

\author{C.-C. Jay Kuo, Min Zhang, Siyang Li, Jiali Duan and Yueru Chen}
\address{University of Southern California, Los Angeles, CA 90089-2564, USA}

\begin{abstract}

The model parameters of convolutional neural networks (CNNs) are
determined by backpropagation (BP).  In this work, we propose an
interpretable feedforward (FF) design without any BP as a reference.
The FF design adopts a data-centric approach. It derives network
parameters of the current layer based on data statistics from the output
of the previous layer in a one-pass manner. To construct convolutional
layers, we develop a new signal transform, called the Saab ({\bf
S}ubspace {\bf a}pproximation with {\bf a}djusted {\bf b}ias) transform.
It is a variant of the principal component analysis (PCA) with an added
bias vector to annihilate activation's nonlinearity.  Multiple Saab
transforms in cascade yield multiple convolutional layers.  As to
fully-connected (FC) layers, we construct them using a cascade of
multi-stage linear least squared regressors (LSRs). The classification
and robustness (against adversarial attacks) performances of BP- and
FF-designed CNNs applied to the MNIST and the CIFAR-10 datasets are
compared.  Finally, we comment on the relationship between BP and FF
designs. 

\begin{keyword}
Interpretable machine learning, convolutional neural networks, principal
component analysis, linear least-squared regression, cross entropy,
dimension reduction. 
\end{keyword}
\end{abstract}

\end{frontmatter}

\section{Introduction}\label{sec:introduction}

Convolutional neural networks (CNNs) have received a lot of attention in
recent years due to their outstanding performance in numerous
applications.  We have also witnessed the rapid development of advanced
CNN models and architectures such as the generative adversarial networks
(GANs) \cite{goodfellow2014generative}, the ResNets \cite{He_2016_CVPR}
and the DenseNet \cite{huang2017densely}. 

A great majority of current CNN literature are application-oriented,
yet efforts are made to build theoretical foundation of CNNs.  Cybenko
\cite{cybenko1989approximation} and Hornik {\em et al.}
\cite{hornik1989multilayer} proved that the multi-layer perceptron (MLP)
is a universal approximator in late 80s.  Recent studies on CNNs
include: visualization of filter responses at various layers
\cite{simonyan2013deep, zeiler2014visualizing, zhou2014object},
scattering networks \cite{mallat2012group, bruna2013invariant,
wiatowski2015mathematical}, tensor analysis \cite{cohen2015expressive},
generative modeling \cite{dai2014generative}, relevance propagation
\cite{bach2015pixel}, Taylor decomposition
\cite{montavon2015explaining}, multi-layer convolutional sparse modeling
\cite{sulam2017multi} and over-parameterized shallow neural network
optimization \cite{soltanolkotabi2018theoretical}.  

More recently, CNN's interpretability has been examined by a few
researchers from various angles. Examples include interpretable
knowledge representations \cite{zhang2017interpretable}, critical nodes
and data routing paths identification \cite{wang2018interpret}, the role
of nonlinear activation \cite{kuo2016understanding}, convolutional
filters as signal transforms \cite{kuo2017cnn,kuo2018data}, etc.
Despite the above-mentioned efforts, it is still challenging to provide
an end-to-end analysis to the working principle of deep CNNs. 

Given a CNN architecture, the determination of network parameters can be
formulated as a non-convex optimization problem and solved by
backpropagation (BP). Yet, since non-convex optimization of deep
networks is mathematically intractable
\cite{soltanolkotabi2018theoretical}, a new methodology is adopted in
this work to tackle the interpretability problem. That is, we propose an
interpretable feedforward (FF) design without any BP and use it as a
reference. The FF design adopts a data-centric approach.  It derives
network parameters of the current layer based on data statistics from
the output of the previous layer in a one-pass manner.  This
complementary methodology not only offers valuable insights into CNN
architectures but also enriches CNN research from the angles of linear
algebra, statistics and signal processing. 

To appreciate this work, a linear algebra viewpoint on machine learning
is essential. An image, its class label and intermediate representations
are all viewed as high-dimensional vectors residing in certain vector
spaces. For example, consider the task of classifying a set of RGB color
images of spatial resolution $32 \times 32$ into $10$ classes. The input
vectors are of dimension $32 \times 32 \times 3=3,072$. Each desired
output is the unit dimensional vector in a 10-dimensional (10D) space.
To map samples from the input space to the output space, we conduct a
sequence of vector space transformations.  Each layer provides one
transformation.  A {\em dimension} of a vector space can have one of
three meanings depending on the context: a {\em representation} unit, a
{\em feature} or a class {\em label}.  The term ``dimension" provides a
unifying framework for the three different concepts in traditional
pattern recognition.  

In our interpretation, each CNN layer corresponds to a vector space
transformation. To take computer vision applications as an example, CNNs
provide a link from the input image/video space to the output decision
space. The output can be an object class (e.g., object classification),
a pixel class (e.g., semantic segmentation) or a pixel value (e.g. depth
estimation, single image super-resolution, etc.) The training data
provide a sample distribution in the input space. We use the input data
distribution to determine a proper transformation to the output space.
The transformation is built upon two well-known ideas: 1) dimension
reduction through subspace approximations and/or projections, and 2)
training sample clustering and remapping.  The former is used in
convolutional layers construction (e.g., filter weights selection and
max pooling) while the latter is adopted to build FC layers. They are
elaborated a little more below. 

The convolutional layers offer a sequence of spatial-spectral filtering
operations. Spatial resolutions become coarser along this process
gradually. To compensate for the loss of spatial resolution, we convert
spatial representations to spectral representations by projecting pixels
in a window onto a set of pre-selected spatial patterns obtained by the
principal component analysis (PCA). The transformation enhances
discriminability of some dimensions since a dimension with a larger
receptive field has a better chance to ``see" more. We develop a new
transform, called the Saab ({\bf S}ubspace {\bf a}pproximation with {\bf
a}djusted {\bf b}ias) transform, in which a bias vector is added to
annihilate nonlinearity of the activation function. The Saab transform
is a variant of PCA, and it contributes to dimension reduction. 

The FC layers provide a sequence of operations that involve ``sample
clustering and high-to-low dimensional mapping". Each dimension of the
output space corresponds to a ground-truth label of a class. To
accommodate intra-class variability, we create sub-classes of finer
granularity and assign pseudo labels to sub-classes.  Consider a
three-level hierarchy -- the feature space, the sub-class space and the
class space. We construct the first FC layer from the feature space to
the sub-class space using a linear least-squared regressor (LSR) guided
by pseudo-labels. For the second FC layer , we treat pseudo-labels as
features and conduct another LSR guided by true labels from the
sub-class space to the class space. A sequence of FC layers actually
corresponds to a multi-layer perceptron (MLP). To the best of our
knowledge, this is the first time to construct an MLP in an FF manner
using multi-stage cascaded LSRs.  The FF design not only reduces
dimensions of intermediate spaces but also increases discriminability of
some dimensions gradually. Through transformations across multiple
layers, it eventually reaches the output space with strong
discriminability. 

LeNet-like networks are chosen to illustrate the FF design methodology
for their simplicity. Examples of LeNet-like networks include the
LeNet-5 \cite{LeNet1998} and the AlexNet \cite{NIPS2012_AlexNet}. They
are often applied to object classification problems such as recognizing
handwritten digits in the MNIST dataset and 1000 object classes in the
ImageNet dataset.  The classification and robustness (against
adversarial attacks) performances of BP- and FF-designed CNNs on the
MNIST and the CIFAR-10 datasets are reported and compared.  It is
important to find a connection between the BP and the FF designs.  To
shed light on their relationship, we measure cross-entropy values at
dimensions of intermediate vector spaces (or layers). 

The rest of this paper is organized as follows. The related background
is reviewed in Sec.  \ref{sec:interpretation}. The FF design of
convolutional layers is described in Sec. \ref{sec:Saab}.  The FF design
of FC layers is presented in Sec.  \ref{sec:l3sr}.  Experimental results
for the MNIST and the CIFAR-10 datasets are given in Sec.
\ref{sec:experiments}.  Follow-up discussion is made in Sec.
\ref{sec:discussion}.  Finally, concluding remarks are drawn in Sec.
\ref{sec:conclusion}. 

\section{Background}\label{sec:interpretation}

\subsection{Computational neuron}

\begin{figure}[htb]
\centering
\includegraphics[width=8cm]{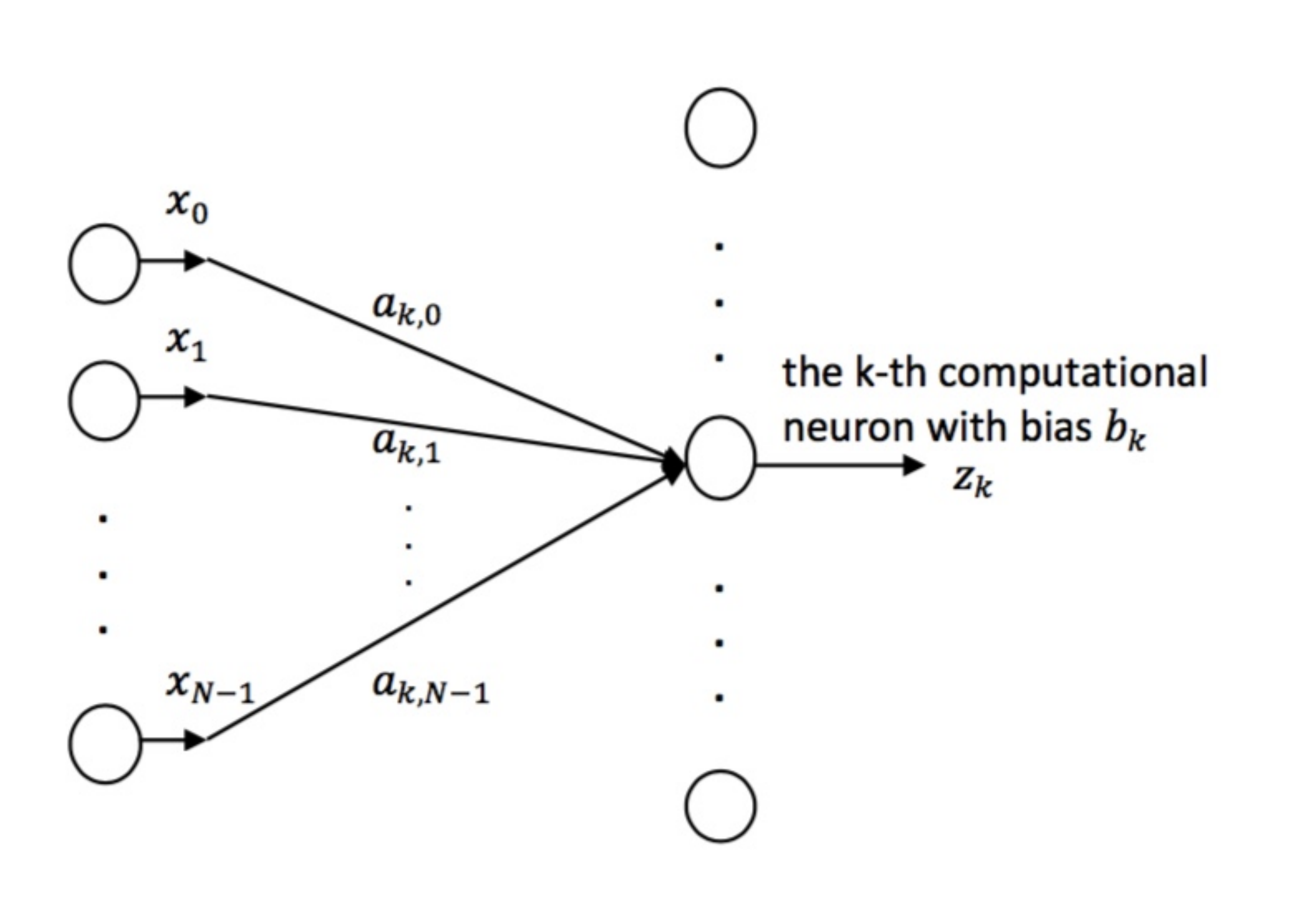}\\(a) \\
\includegraphics[width=12cm]{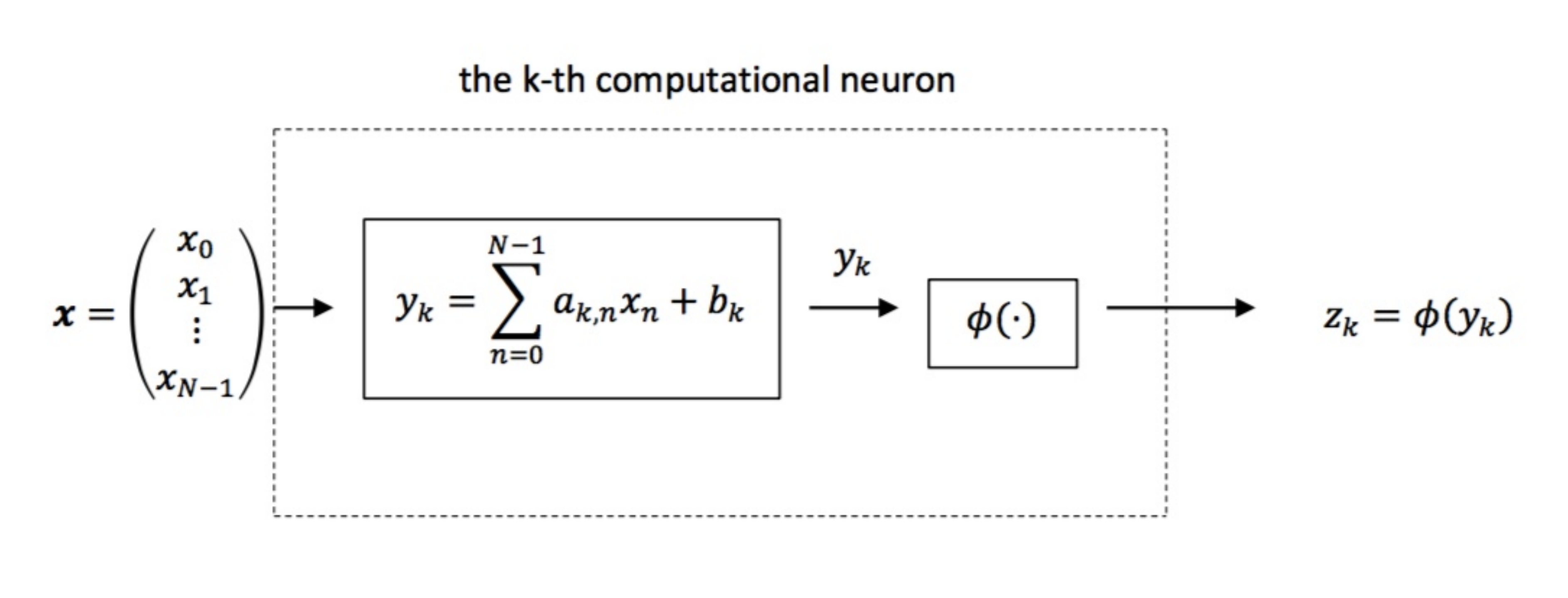}\\(b)
\caption{Illustration of a computational neuron: (a) the input-output 
connection, and (b) the block diagram inside one neuron.}\label{fig:neuron}
\end{figure}

The computational neuron serves as the basic building element of CNNs.
As shown in Fig. \ref{fig:neuron}, it consists of two stages: 1) affine
computation and 2) nonlinear activation.  The input is an
$N$-dimensional random vector ${\bf x}=(x_0, x_1, ... , x_{N-1})^T$. The
$k$th neuron has $N$ filter weights that can be expressed in vector form
as ${\bf a}_k=(a_{k,0}, a_{k,1}, \cdots , a_{k,N-1})^T$, and one bias
term $b_k$.  The affine computation is
\begin{equation}\label{eq:affine}
y_k = \sum_{n=0}^{N-1} a_{k,n} x_n + b_k = {\bf a}_k^T {\bf x} + b_k, 
\quad k=0, 1, \cdots, K-1,
\end{equation}
where ${\bf a}_k$ is the filter weight vector associated with the $k$th
neuron. With the ReLU nonlinear activation function, the output after
ReLU can be written as
\begin{equation}\label{eq:na}
z_k= \phi (y_k) = \max (0, y_k).
\end{equation}

In the following discussion, we assume that input ${\bf x}$ is a
zero-mean random vector and the set of filter weight vectors is
normalized to to be with unit length ({\em i.e.} $||{\bf a}_k||=1$).
The filter weight vectors are adjustable in the training yet they are
given and fixed in the testing. To differentiate the two situations, we
call them anchor vectors in the testing case.  Efforts have been made to
explain the roles played by anchor vector ${\bf a}_k$ and nonlinear
activation $\phi(\cdot)$ in \cite{kuo2016understanding, kuo2017cnn,
kuo2018data}. 

\subsection{Linear space spanned by anchor vectors}

It is easier to explain the role of anchor vectors by setting the bias
term to zero. This constraint will be removed in Sec. \ref{sec:Saab}.
If $b_k=0$, Eq. (\ref{eq:affine}) reduces to $y_k = {\bf a}_k^T {\bf
x}$. Suppose that there are $K$ anchor vectors with $k=0, 1, \cdots,
K-1$ in a convolutional layer. One can examine the role of each anchor
vector individually and all anchor vectors jointly. They lead to two
different interpretations. 
\begin{enumerate}
\item A set of parallel correlators \cite{kuo2016understanding,
kuo2017cnn} \\
If the correlation between ${\bf a}_k$ and ${\bf x}$ is weak (or
strong), response $y_k$ will have a small (or large) magnitude.  Thus,
each anchor vector can be viewed as a correlator or a matched filter.
We can use a set of correlators to extract pre-selected patterns from
input ${\bf x}$ by thresholding $|y_k|$. These correlators can be
determined by BP. In a FF design, one can apply the $k$-means
algorithm to input samples to obtain $K$ clusters and set their
centroids to ${\bf a}_k$. 
\item A set of unit vectors that spans a linear subspace
\cite{kuo2018data} \\
By considering the following set of equations jointly:
\begin{equation}\label{eq:projection}
y_k = {\bf a}_k^T {\bf x}, \quad k=0, 1, \cdots K-1,
\end{equation}
the output vector, ${\bf y}=(y_0, y_1, \cdots , y_{K-1})^T$, is the
projection of the input vector ${\bf x}$ onto a subspace spanned by
${\bf a}_k$, $k=0, 1, \cdots, K-1$.  One can get an approximate to ${\bf
x}$ in the space spanned by anchor vectors from the projected output
${\bf y}$. With this interpretation, we can use the principal component
analysis (PCA) to find a subspace and determine the anchor vector set
accordingly. 
\end{enumerate}

The above two interpretations allow us to avoid the BP training
procedure for filter weights in convolutional layers. Instead, we can
derive anchor vectors from the statistics of input data. That is, we can
compute the covariance matrix of input vectors ${\bf x}$ and use the
eigenvectors as the desired anchor vectors ${\bf a}_k$.  This is a
data-centric approach. It is different from the traditional BP approach
that is built upon the optimization of a cost function defined at the
system output. 

\subsection{Role of nonlinear activation}

The need of nonlinear activation was first explained in
\cite{kuo2016understanding}. The main result is summarized below.
Consider two inputs ${\bf x}_1={\bf x}$ and ${\bf x}_2=-{\bf x}$ with
the simplifying assumption:
\begin{equation}\label{eq:simplifying}
{\bf a}_k^T {\bf x} \approx 1, \mbox{   and   } 
{\bf a}_{k'}^T {\bf x} \approx 0 \mbox{ if } k' \ne k.
\end{equation}
The two inputs, ${\bf x}_1$ and ${\bf x}_2$, are negatively correlated.
For example, if ${\bf x}_1$ is a pattern composed by three vertical
stripes with the middle one in black and two side ones in white, then
${\bf x}_2$ is also a three-vertical-stripe pattern with the middle one
in white and two side ones in black. They are different patterns, yet
one can be confused for the other if there is no ReLU. To show this, we
first compute their corresponding outputs
\begin{eqnarray}\label{eq:outputs}
{\bf y}_1^T & = & ({\bf a}_1^T, \cdots, {\bf a}_k^T, \cdots, 
{\bf a}_K^T){\bf x}_1 \approx (0, \cdots, 0, 1, 0, \cdots, 0),  \\
{\bf y}_2^T & = & ({\bf a}_1^T, \cdots, {\bf a}_k^T, \cdots, 
{\bf a}_K^T){\bf x}_2 \approx (0, \cdots, 0, -1, 0, \cdots, 0),
\end{eqnarray}
where $1$ and $-1$ appear in the $k$th element as shown above.  Vectors
${\bf y}_1$ and ${\bf y}_2$ will serve as inputs to the next
stage. The outgoing links from the $k$th node can take positive or
negative weights. If there is no ReLU, a node at the second layer cannot
differentiate whether the input is ${\bf x}_1$ or ${\bf x}_2$ since the
following two situations yield the same output: \vspace{-1ex}
\begin{itemize}
\setlength{\itemsep}{-2pt}
\item[(a)] a positive correlation (${\bf y}_1$) followed by a positive
outgoing link from node $k$; and
\item[(b)] a negative correlation (${\bf y}_2$) followed by a negative
outgoing link from node $k$. 
\end{itemize}
\vspace{-1ex}
Similarly, the following two situations will also yield the same output:
\vspace{-1ex}
\begin{itemize}
\setlength{\itemsep}{-2pt}
\item[(a)] a positive correlation (${\bf y}_1$) followed by a negative
outgoing link from node $k$; and
\item[(b)] a negative correlation (${\bf y}_2$) followed by a positive
outgoing link from node $k$. 
\end{itemize}
This is called the sign confusion problem.  The ReLU operator plays the
role of a rectifier that eliminates case (b). In the example, input
${\bf x}_2$ will be blocked by the system. A trained CNN has its
preference on images over their foreground/background reversed ones.  A
classification example conducted on the original and reversed MNIST
datasets was given in \cite{kuo2016understanding} to illustrate this
point. 

To resolve the sign confusion problem, a PCA variant called the Saak
transform was proposed in \cite{kuo2018data}. The Saak transform
augments transform kernel ${\bf a}_k$ with its negative $-{\bf a}_k$,
leading to $2K$ transform kernels in total. Any input vector, ${\bf x}$,
will have a positive/negative correlation pair with kernel pair $({\bf
a}_k,-{\bf a}_k)$. When a correlation is followed by ReLU, one of the
two will go through while the other will be blocked. In other words, the
Saak transform splits positive/negative correlations, ${\bf y}_k={\bf
a}_k^T {\bf x}$, into positive/negative two channels.  To resolve the
sign confusion problem, it pays the price of doubled spectral
dimensions. In the following section, we will introduce another PCA
variant called the Saab (Subspace approximation with adjusted bias)
transform. The Saab transform can address the sign confusion problem and
avoid the spectral dimension doubling problem at the same time. 

\begin{figure}[htb]
\centering
\includegraphics[width=12cm]{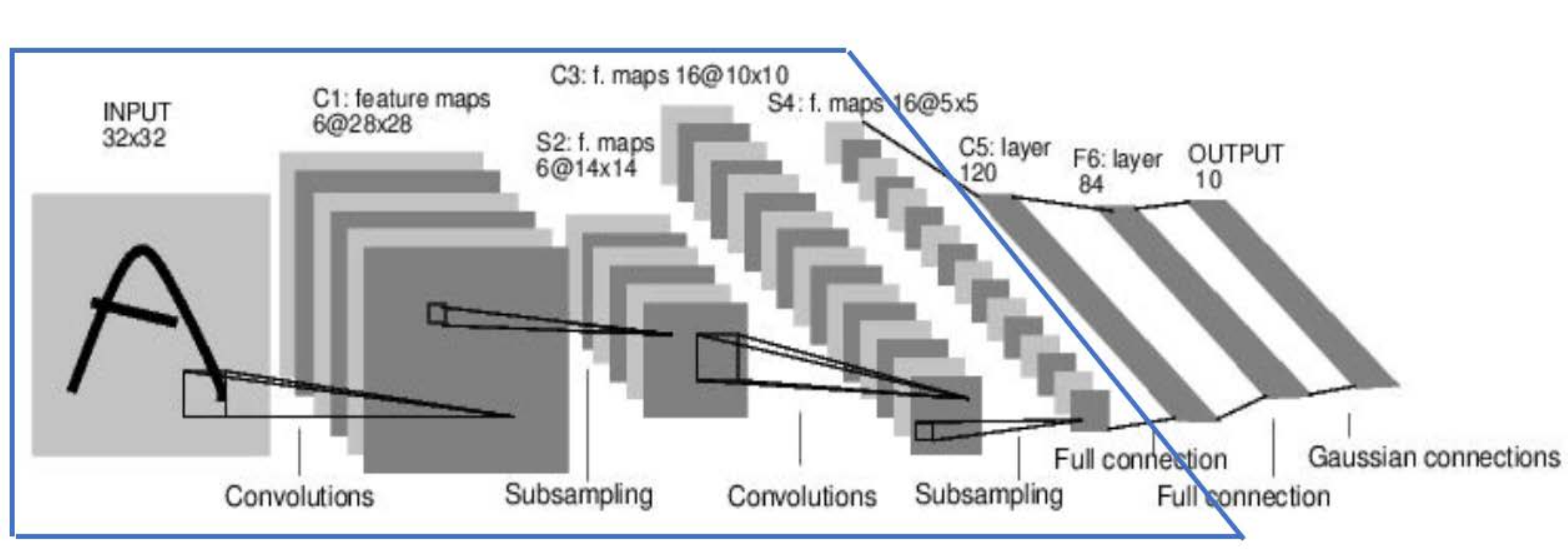}
\caption{The LeNet-5 architecture \cite{LeNet1998}, where the
convolutional layers are enclosed by a blue parallelogram.}
\label{fig:LeNet-5-Conv}
\end{figure}

\section{Feedforward design of convolutional layers}\label{sec:Saab}

In this section, we study the construction of convolutional layers in
the LeNet-5 as shown in the enclosed parallelogram in Fig.
\ref{fig:LeNet-5-Conv}. We first examine the spatial-spectral filtering
in Sec. \ref{subsec:filtering}. The Saab transform and its bias
selection is studied in Sec. \ref{subsec:bias}. Then, we discuss the max
pooling operation in Sec. \ref{subsec:pooling}.  Finally, we comment on
the effect of cascaded convolutional layers in Sec.
\ref{subsec:compound}. 

\subsection{Spatial-spectral filtering}\label{subsec:filtering}

There is a clear distinction between representations and features in
traditional image processing. Image representations are obtained by
transforms such as the Fourier transform, the discrete Cosine transform
and the wavelet transform, etc. Image transforms are invertible.  Image
features are extracted by detectors such as those for edges, textures
and salient points. The distinction between image representations and
image features is blurred in CNNs as mentioned in Sec.
\ref{sec:introduction}. An input image goes through a sequence of vector
space transformations. Each dimension of intermediate spaces can be
viewed as either a representation or a feature.  The cascade of
spatial-spectral filtering and spatial pooling at a convolutional layer
provides an effective way in extracting discriminant dimensions.
Without loss of generality, we use the LeNet-5 applied to the MNIST
dataset as an illustrative example. 

It is not a good idea to conduct pixel-wise image comparison to
recognize handwritten digits since there exists a wide range of
varieties in the spatial domain for the same digit. Besides, the
pixel-wise comparison operation is sensitive to small translation and
rotation. For example, if we shift the same handwritten digit image ``1"
horizontally by several pixels, the original and shifted images will
have poor match in the spatial domain. A better way is to consider the
neighborhood of a pixel, and find a spectral representation for the
neighborhood.  For example, we use a patch of size $5 \times 5$ centered
at a target pixel as its neighborhood. The stroke inside a patch is a 2D
pattern. We can use PCA to find a set of dominant stroke patterns
(called anchor vectors) to form a vector space and represent any
neighborhood pattern as a linear combination of anchor vectors. 

In the multi-stage design, the convolutional layers provide a sequence
of spatial-spectral transformations that convert an input image to its
joint spatial-spectral representations layer by layer. Spatial
resolutions become lower gradually. To compensate the spatial resolution
loss, we trade spatial representations for spectral representations by
projecting local spatial-spectral cuboids onto PCA-based kernels.  The
main purpose is to enhance discriminability of some dimensions.
Generally, a spatial-spectral component with a larger receptive field
has a better chance to be discriminant since it can ``see" more. Another
advantage of the PCA-based subspace approximation is that it does not
demand image labels. 

One related question is whether to conduct the transform in overlapping or non-overlapping windows. Signal transforms are often 
conducted in non-overlapping windows for computational and storage 
efficiency. For example, the block discrete Cosine transform (DCT) is 
adopted in image/video compression.  For the same reason, the Saak 
transform in \cite{kuo2018data} is conducted on non-overlapping 
windows.However, the LeNet-5 uses overlapping windows. The MNIST dataset
contains gray-scale images of dimension $32 \times 32$.  At the first
convolutional layer, the LeNet-5 has 6 filters of size $5\times 5$ with
stride equal to one. The output image cuboid has a dimension of $28
\times 28 \times 6$ by taking the boundary effect into account.  If we
ignore the boundary effect and apply $K$ spatial-spectral filtering
operations at every pixel, the output data dimension is enlarged by a
factor of $K$ with respect to the input.  This redundant representation
seems to be expensive in terms of higher computational and storage
resources.  However, it has one advantage. That is, it provides a {\em
richer} feature set for selection. Redundancy is controlled by the
stride parameter in CNN architecture specification.

\subsection{Saab transform and bias selection}\label{subsec:bias}

We repeat the affine transform in Eq. (\ref{eq:affine}) below:
\begin{equation}\label{eq:affine_r}
y_k = \sum_{n=0}^{N-1} a_{k,n} x_n + b_k = {\bf a}_k^T {\bf x} 
+ b_k, \quad k=0, 1, \cdots, K-1,
\end{equation}
The Saab transform is nothing but a specific way in selecting anchor
vector ${\bf a}_k$ and bias term $b_k$. They are elaborated in this
subsection. 

{\bf Anchor vectors selection.} By following the treatment in
\cite{kuo2016understanding, kuo2017cnn, kuo2018data}, we first set
$b_k=0$ and divide anchor vectors into two categories:
\begin{itemize}
\item DC anchor vector ${\bf a}_0 = \frac{1}{\sqrt{N}} (1, \cdots, 1)^T$.
\item AC anchor vectors ${\bf a}_k$, $k=1, \cdots K-1$.
\end{itemize}
The terms ``DC" and ``AC" are borrowed from circuit theory, and they
denote the ``direct current" and the ``alternating current",
respectively. Based on the two categories of anchor vectors, we
decompose the input vector space, $\mathcal{S}=R^N$, into the direct sum
of two subspaces:
\begin{equation}\label{eq:space_decomposition}
\mathcal{S}=\mathcal{S}_{DC} \oplus \mathcal{S}_{AC},
\end{equation}
where $\mathcal{S}_{DC}$ is the subspace spanned by the DC anchor and
and $\mathcal{S}_AC$ is the subspace spanned by the AC anchors.  They
are called the DC and AC subspaces accordingly.  For any vector ${\bf x}
\in R^N$, we can project ${\bf x}$ to ${\bf a}_0$ to get its DC
component. That is, we have
\begin{equation}\label{eq:mean}
{\bf x}_{DC}={\bf x}^T {\bf a}_0 = \frac{1}{\sqrt{N}} \sum_{n=0}^{N-1} x_n.
\end{equation}
Subspace $\mathcal{S}_{AC}$ is the orthogonal complement to
$\mathcal{S}_{DC}$ in $\mathcal{S}$. We can express the AC component of
${\bf x}$ as
\begin{equation}\label{eq:AC}
{\bf x}_{AC}={\bf x} - {\bf x}_{DC}.
\end{equation}
Clearly, we have ${\bf x}_{DC} \in \mathcal{S}_{DC}$ and ${\bf x}_{AC}
\in \mathcal{S}_{AC}$. We conduct the PCA on all possible ${\bf x}_{AC}$
and, then, choose the first $(K-1)$ principal components as AC anchor
vectors ${\bf a}_k$, $k=1, \cdots K-1$. 

{\bf Bias selection.} Each bias term, $b_k$, in Eq.
(\ref{eq:affine_r}), provides one extra degree of freedom per neuron for
the end-to-end penalty minimization in the BP design. Since it plays no
role in linear subspace approximation, it was ignored in
\cite{kuo2016understanding, kuo2017cnn, kuo2018data}.  Here, the bias
term is leveraged to overcome the sign confusion problem in the Saab
transform. We impose two constraints on the bias terms. 
\begin{itemize}
\item[(B1)] Positive response constraint \\
We choose the $k$th bias, $b_k$, to guarantee the $k$th response a 
non-negative value. Mathematically, we have
\begin{equation}\label{eq:affine2}
y_k=\sum_{n=0}^{N-1}a_{k,n}x_n+b_k={\bf a}_k^T {\bf x}+b_k \ge 0,
\end{equation}
for all input ${\bf x}$.
\item[(B2)] Constant bias constraint \\
We demand that all bias terms are equal; namely,
\begin{equation}\label{eq:b_constant}
b_0=b_1= \cdots = b_{K-1} \equiv d \sqrt{K}. 
\end{equation}
\end{itemize}

Due to Constraint (B1), we have
\begin{equation}\label{eq:na2}
z_k= \phi (y_k) = \max (0, y_k) = y_k.
\end{equation}
That is, we have the same output with or without the ReLU operation.
This simplifies our CNN analysis greatly since we remove nonlinearity
introduced by the activation function. Under Constraint (B2), we only
need to determine a single bias value $d$ (rather than $K$ different
bias values). It makes the Saab transform design easier. Yet, there is a
deeper meaning in (B2). 

Let ${\bf x}=(x_0, \cdots, x_{N-1})$ lie in the AC subspace of the input 
space, $R^N$. We can re-write Eq.  (\ref{eq:affine_r}) in vector form as
\begin{equation}\label{eq:vector}
{\bf y} = {\bf x}^T {\bf a}_0 + \sum_{k=1}^{K-1} {\bf x}^T {\bf a}_k +  
d \sqrt{K} {\bf l}, 
\end{equation}
where ${\bf l}=\frac{1}{\sqrt{K}}(1, \cdots, 1)^T$ is the unit
constant-element vector in the output space $R^K$. The first term in Eq.
(\ref{eq:vector}) is zero due to the assumption. The second and third
terms lie in the AC and DC subspaces of the output space, $R^K$,
respectively.  In other words, the introduction of the constant-element
bias vector has no impact on the AC subspace but the DC subspace when
multiple affine transforms are in cascade.  It is essential to impose
(B2) so that the multi-stage Saab transforms in cascade are
mathematically tractable. That is, the constant-element bias vector
${\bf b}=d {\bf 1}$ always lies in the DC subspace, $\mathcal{S}_{DC}$.
It is completely decoupled from $\mathcal{S}_{AC}$ and its PCA.  We
conduct PCA on the AC subspace, $\mathcal{S}_{AC}$, layer by layer to
obtain AC anchor vectors at each layer. 

The addition of a constant-element bias vector to the transformed output
vector is nothing but shift each output response by a constant amount.
We derive a lower bound on this amount in the appendix.  The bias
selection rule can be stated below.
\begin{itemize}
\item {\bf Bias selection rule} \\
All bias terms should be equal (i.e., $b_0=b_1= \cdots = b_{K-1}$), 
and they should meet the following constraint:
\begin{equation}\label{eq:d3}
b_k \ge \max_{\bf x} || {\bf x} ||, \quad k=0, \cdots, K-1.
\end{equation}
where ${\bf x} \in R^N$ is an input vector and $K$ is the dimension 
of the output space $R^K$. 
\end{itemize}

\subsection{Spatial pooling}\label{subsec:pooling} 

\begin{figure}[htb]
\centering
\includegraphics[width=6cm]{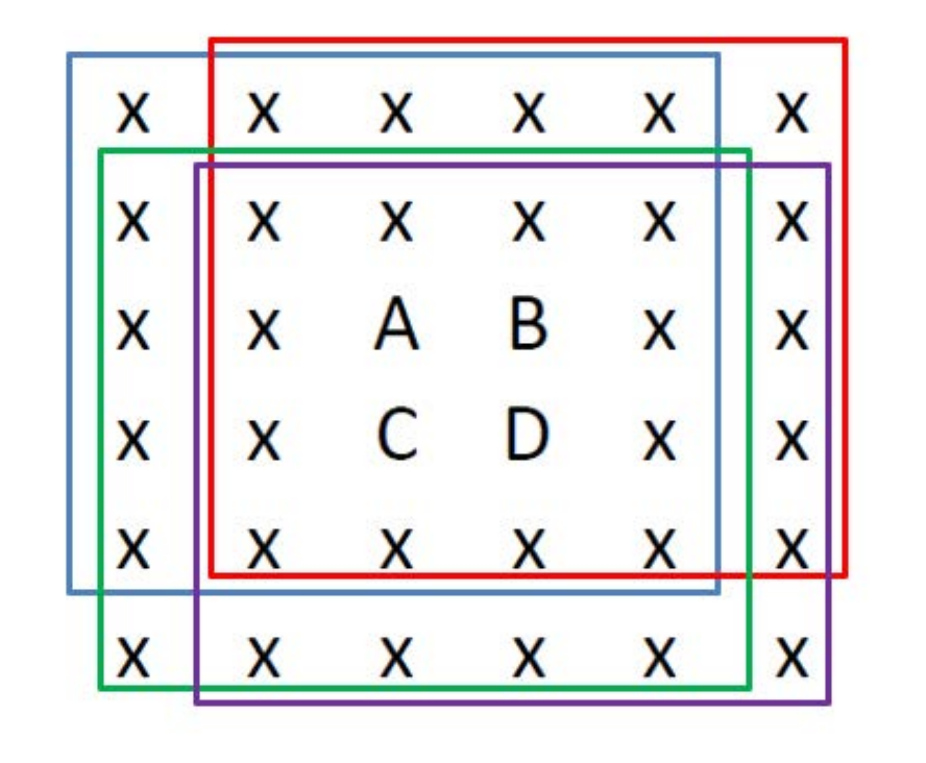}
\caption{Illustration of the spatial pooling, where A, B, C, D denote four
spatial locations within a $2 \times 2$ block where the maximum pooling
operation is conducted at each spectral component independently.}\label{fig:pooling}
\end{figure}

{\bf Rationale of maximum pooling.} Spatial pooling helps reduce
computational and storage resources.  However, this cannot explain a
common observation: ``Why maximum pooling outperforms the average
pooling?" Here, we interpret the pooling as another filtering operation
that preserves significant patterns and filters out insignificant ones.
Fig.  \ref{fig:pooling} shows four spatial locations, denoted by A, B,
C, D, in a representative $2 \times 2$ non-overlapping block. Suppose
that $K$ filters are used to generate responses at each location.  Then,
we have a joint spatial-spectral response vector of dimension $4 \times
K$.  Pooling is used to reduce the spatial dimension of this response
vector from $4$ to $1$ so that the new response vector is of dimension
$1 \times K$.  It is performed at each spectral component independently.
Furthermore, all response values are non-negative due to the ReLU
function. 

Instead of viewing convolution and pooling operations as two individual
ones, we examine them as a whole.  The new input is the union of the
four neighborhoods, which is a patch of size $36=6 \times 6$. The
compound operations, consisting of convolution and pooling, is to
project the enlarged neighborhood patch of dimension $36$ to a subspace
of dimension $K$ spanned by $K$ anchor vectors (or filters).  The
convolution-plus-pooling operations is equivalent to:
\begin{enumerate}
\item Projecting a patch of size $36=6 \times 6$ to any smaller one 
of size $5 \times 5$ centered at locations A-D with zero padding in
uncovered areas;
\item Projecting the patch of size $5 \times 5$ to all anchor vectors. 
\end{enumerate}
Each spectral component corresponds to a visual pattern. For a given
pattern, we have four projected values obtained at locations A, B, C, D.
By maximum pooling, we choose the maximum value among the four.  That
is, we search the target visual pattern in a slightly larger window
(i.e. of size $6 \times 6$) and use the maximum response value to
indicate the best match within this larger window. 

On the other hand, infrequent visual patterns that are less relevant to
the target task will be suppressed by the compound operation of
spatial-spectral filtering (i.e.  the Saab transform) and pooling. The
convolutional kernels of the Saab transform are derived from the PCA.
The projection of these patterns on anchor vectors tend to generate
small response values and they will be removed after pooling. 

The same principle can be generalized to pooling in deeper layers. A
spatial location in a deep layer corresponds to a receptive field in the
input source image.  The deeper the layer the larger the receptive
field. A spectral component at a spatial location can be interpreted as
a projection to a dominant visual pattern inside its receptive field.  A
$2 \times 2$ block in a deep layer correspond to the union of four
overlapping receptive fields, each of which is associated with one
spatial location.  To summarize, the cascade of spatial-spectral
filtering and maximum pooling can capture ``visually similar but
spatially displaced" patterns. 

\begin{figure}[htb]
\centering
\includegraphics[width=12cm]{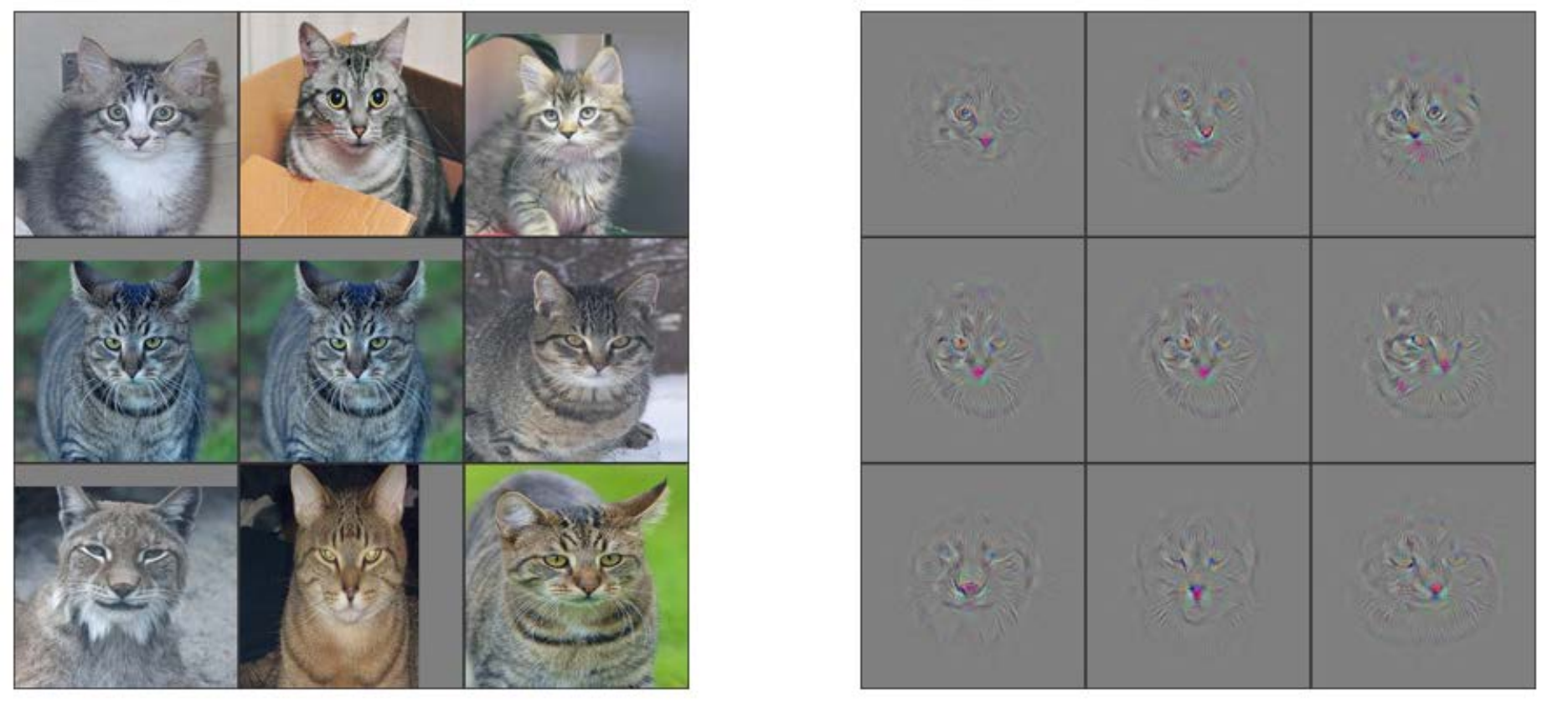}
\caption{Left: Display of nine top-ranked images that have the strongest
responses with respect to a certain convolutional filter (called the cat
face filter) in the 5th convolutional layer of the AlexNet. Right: Their
corresponding filter responses.}\label{fig:cat-face}
\end{figure}

\subsection{Multi-layer compound filtering}\label{subsec:compound} 

\begin{figure}[htb]
\centering
\includegraphics[width=12cm]{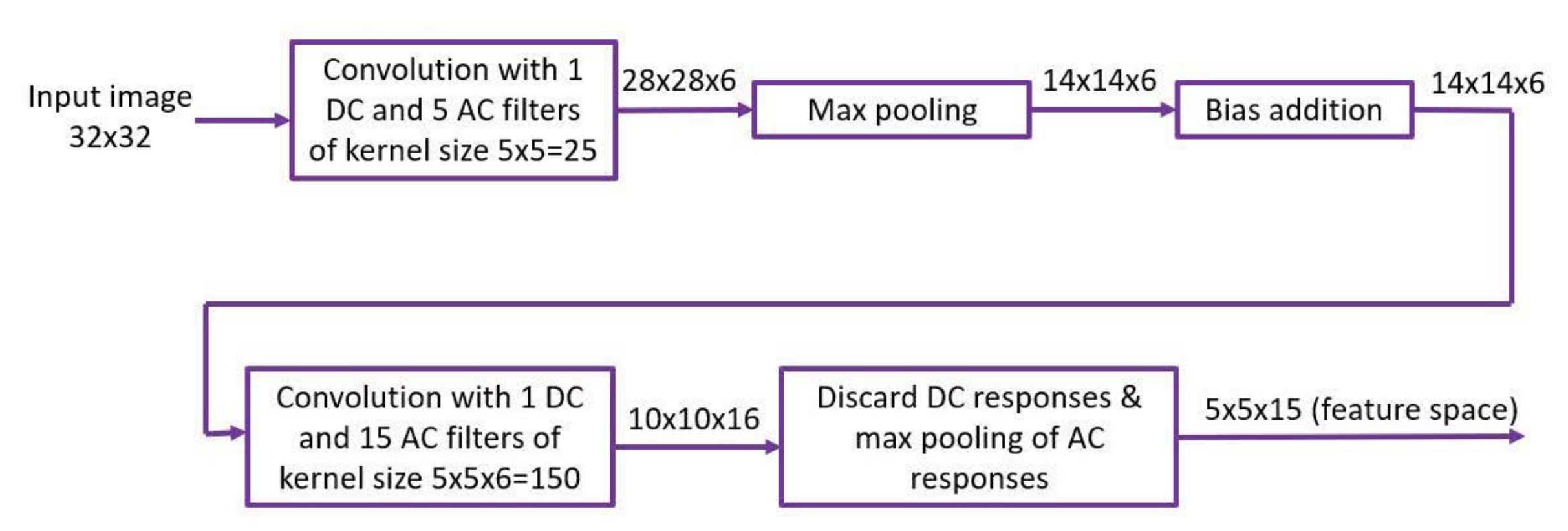}
\caption{Summary of the FF design of the first two convolutional 
layers of the LeNet-5.}\label{fig:ff-block-diagram}
\end{figure}

The cascade of multiple convolutional layers can generate a rich set of
image patterns of various scales as object signatures. We call them
compound filters. One can obtain interesting compound filters through
the BP design. To give an example, we show the nine top-ranked images
that have the strongest responses with respect to a certain
convolutional filter (called the cat face filter) in the 5th
convolutional layer of the AlexNet \cite{NIPS2012_AlexNet} and their
corresponding filter responses in Fig. \ref{fig:cat-face}. We see cat
face contours clearly in the right subfigure.  They are of size around
$100 \times 100$. The compound filtering effect is difficult to
implement using a single convolutional layer (or a single-scale
dictionary).  In the FF design, a target pattern is typically
represented as a linear combination of responses of a set of orthogonal
PCA filters.  These responses are signatures of the corresponding
receptive field in the input image.  There is no need to add the bias
term in the last convolutional layer since a different design
methodology is adopted for the construction of FC layers.  The
block-diagram of the FF design of the first two convolutional layers of
the LeNet-5 is shown in Fig.  \ref{fig:ff-block-diagram}. 

\section{Feedforward design of FC layers}\label{sec:l3sr}

\begin{figure}[htb]
\centering
\includegraphics[width=12cm]{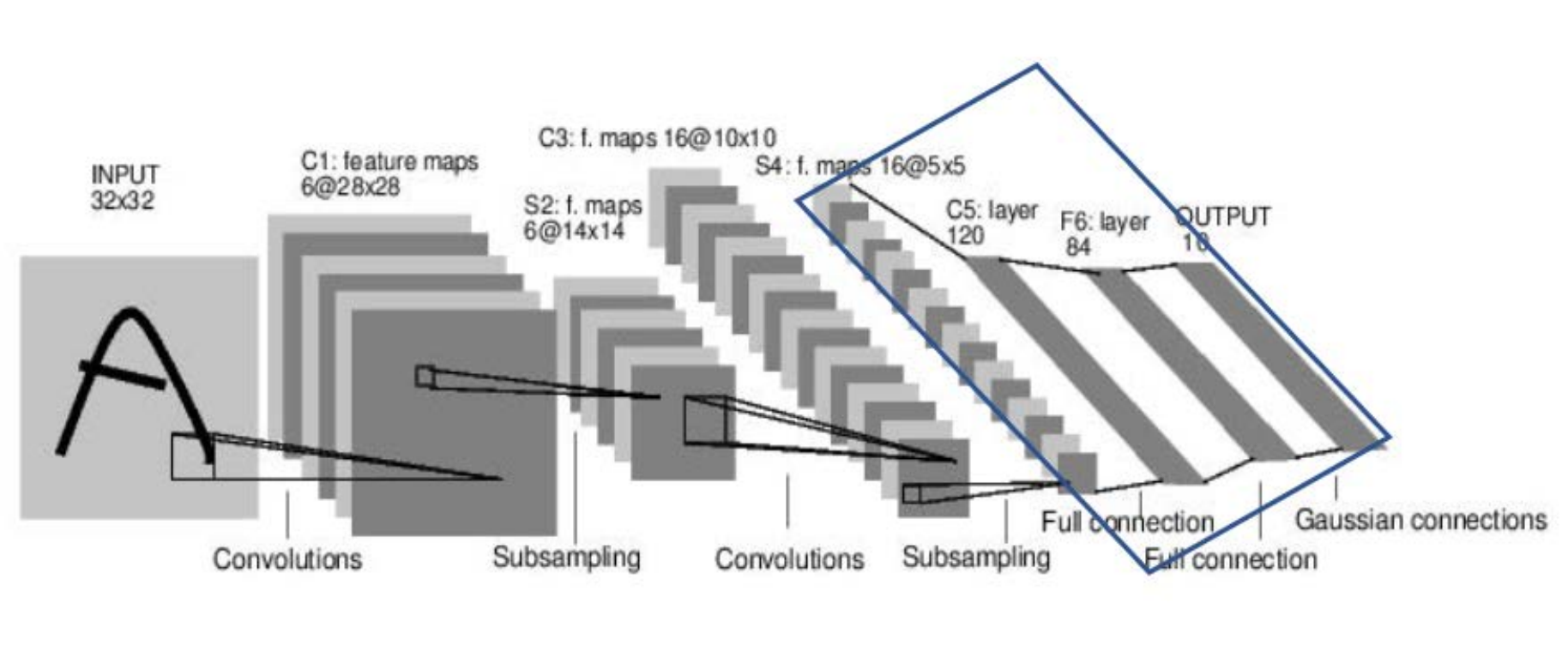}
\caption{The LeNet-5 architecture \cite{LeNet1998}, where the FC 
layers are enclosed by a blue parallelogram.}\label{fig:LeNet-FC}
\end{figure}

The network architecture the LeNet-5 is shown in Fig.
\ref{fig:LeNet-FC}, where FC layers enclosed by a blue parallelogram.
The input to the first FC layer consists of data cuboids of dimension
$5\times5\times 16$ indicated by S4 in the figure. The output layer
contains 10 output nodes, which can be expressed as a 10-dimensional
vector. There are two hidden layers between the input and the output of
dimensions 120 and 84, respectively.  We show how to construct FC layers
using a sequence of label-guided linear least-squared regressors. 

\subsection{Least-squared regressor (LSR)}\label{subsec:l2sr}

In the FF design, each FC layer is treated as a linear least-squared
regressor.  The output of each FC layer is a one-hot vector whose
elements are all set to zero except for one element whose value is set
to one.  The one-hot vector is typically adopted in the output layer of
CNNs. Here, we generalize the concept and use it at the output of each
FC layer.  There is one challenge in this generalization.  That is,
there is no label associated with an input vector when the output is one
of the hidden layers. To address it, we conduct the k-means clustering
on the input, and group them into $Q$ clusters where $Q$ is the number
of output nodes.  Then, each input has two labels -- the original class
label and the new cluster label.  We combine the class and the cluster
labels to create $K$ pseudo-labels for each input. Then, we can set up a
linear least-squared regression problem using the one-hot vector defined
by $K$ pseudo-labels at this layer.  Then, we can conduct the
label-guided linear least-squared regression in multiple stages (or
multi-layers). 

To derive a linear least-squared regressor, we set up a set of
linear equations that relates the input and the output variables.
Suppose that ${\bf x}=(x_1, x_2, \cdots, x_n)^T \in R^n$ and ${\bf
y}=(y_1, y_2, \cdots, y_c)^T \in R^c$ are input and output vectors.
That is, we have
\begin{equation}\label{eq:l3sr}
\left[
\begin{array}{ccccc}
a_{11} & a_{12} & \cdots & a_{1n} & w_1 \\
a_{21} & a_{22} & \cdots & a_{2n} & w_2 \\
\vdots & \vdots & \ddots & \vdots & \vdots \\
a_{c1} & a_{c2} & \cdots & a_{cn} & w_c
\end{array}
\right]
\left[\begin{array}{c}
x_{1} \\
x_{2} \\
\vdots \\
x_{n} \\
1
\end{array}
\right]
=
\left[\begin{array}{c}
y_{1} \\
y_{2} \\
\vdots \\
y_{c} \\
\end{array}
\right],
\end{equation}
where $w_1$, $w_2$, $\cdots$, $w_c$ are scalars to account for $c$ bias
terms.  After nonlinear activation, each FC layer is a rectified linear
least-squared regressor. 

There are three FC layers in cascade in the LeNet-5. 
\begin{itemize}
\item First FC layer (or Stage I): $n_1=375$ and $m_1=120$.
\item Second FC layer (or Stage II): $n_2=120$ and $m_2=84$.
\item Third FC layer (or Stage III): $n_3=84$ and $m_3=10$.
\end{itemize}
The input data to the first FC layer is a data cuboid of dimension $5
\times 5 \times 15=375$ since the DC responses are removed (or dropped
out). The output data of the third FC layer is the following ten one-hot
vector of dimension $10$:
\begin{equation}
(1, 0, \cdots, 0)^T, (0, 1, 0, \cdots, 0)^T, \cdots, 
(0, \cdots, 0, 1, 0)^T, (0, \cdots, 0, 1)^T.
\end{equation}
Each one-hot vector denotes the label of a hand-written digit. For
example, we can use the above ten one-hot vector to denote digits ``0",
``1", $\cdots$, ``9", respectively. 

By removing the two middle hidden layers of dimensions 120 and 84 and
conducting the linear least-squared regression on the input feature
vector of dimension 375 with 10 one-hot vectors as the desired output
directly, this is nothing but traditional one-stage least-squared
regression. However, its performance is not very good. The advantage of
introducing hidden layers and the way to generate more pseudo-labels
will be discussed below. 

\begin{figure}[htb]
\centering
\includegraphics[width=12cm]{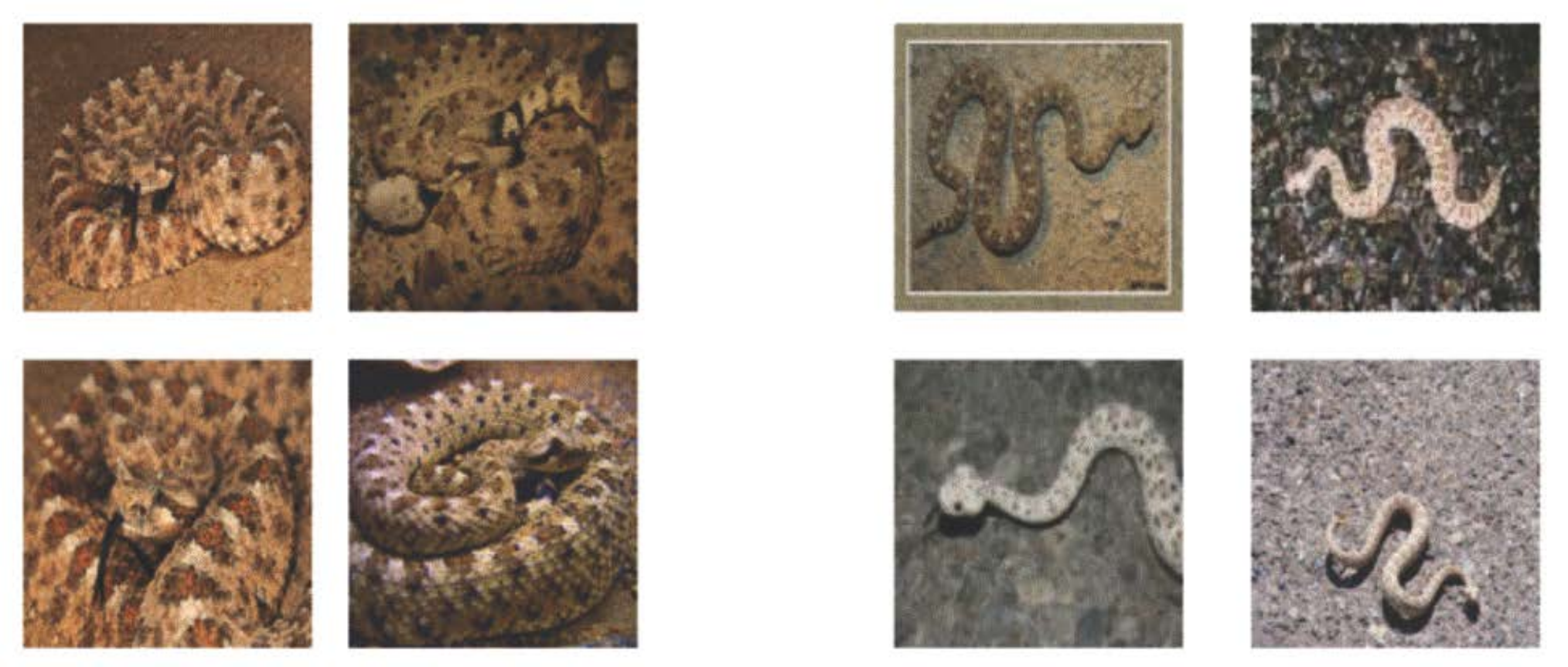}
\caption{Illustration of intra-class variabilities \cite{kuo2017cnn}.}
\label{fig:block-diagram}
\end{figure}

\subsection{Pseudo-labels generation}\label{subsec:pseudo}

To build least-squared regressors in Stages I and II, we need to define
labels for input. Here, we consider the combination of two factors: the
original label (denoted by $0, 1, \cdots, 9$) and the auxiliary label
(denoted by i, ii, $\cdots$, x, xi, xii, etc.) of an input data sample.
Each input training sample has its own original label.  To create its
auxiliary label, we conduct the k-means clustering algorithm on training
samples of the same class. For example, we can divide samples of the
same digit into 12 clusters to generate 12 pseudo classes. The centroid
of a class provides a representative sample for that class. The reason
to generate pseudo classes is to capture the diversity of a single class
with more representative samples, say, from one to twelve in the
LeNet-5. 

As a result, we can set up 120 linear equations to map samples in the
input feature space to 120 one-hot vectors, which form the output
feature space. A one-hot vector is a unit vector in an orthogonal
feature space. Ideally, an LSR is constructed to map all samples in a
pseudo-class to the unit vector of the target dimension and block the
mapping of these samples to other unit vectors.  This is difficult to
achieve in practice since its performance is highly correlated with the
feature distribution in the input space. If two pseudo classes have
strong overlaps in the feature space (say, some 7's and 9's are visually
similar), we expect a significant projection of samples in one pseudo
class to the one-hot vector of another pseudo-class, vice versa. This is
called ``leakage" (with respect to the original pseudo class) or
``interference" (with respect to other pseudo classes). 

The output of the last convolutional layer has a physical explanation in
the FF design. It is a spatial-spectral transform of an input image. It
provides a feature vector for the classifier in later layers.  Once it
is fed to the FC layer, a mathematical model is used to align the input
and output feature spaces by mapping samples from a pseudo class to one
of orthogonal unit vectors that span the output feature space.  The
purpose of feature space alignment is to lower the cross-entropy value
and produce more discriminant features.  Thus, the cascade of LSRs
conducted in multiple FC layers is nothing but a sequential feature
space alignment procedure. In contrast, BP-designed CNNs do not have
such a clear cut in roles played by the convolutional and the FC layers. 

\section{Experimental Results}\label{sec:experiments}

We will show experimental results conducted on two popular datasets: the
MNIST dataset\footnote{http://yann.lecun.com/exdb/mnist/} and the
CIFAR-10 dataset \footnote{https://www.cs.toronto.edu/~kriz/cifar.html}. 
Our implementation codes are available in the GitHub website.

{\bf Network architectures.} The LeNet-5 architecture targets at
gray-scale images only. Since the CIFAR-10 dataset is a color image
dataset, we need to modify the network architecture slightly. The
parameters of the original and the modified LeNet-5 are compared in
Table \ref{table:mLeNet-5}. Note that the modified LeNet-5 keeps the
architecture of two convolutional layers and three FC layers, which
include the last output layer. The modification is needed since the
input images are color images in the CIFAR-10 dataset. We use more
filters at all layers, which are chosen heuristically. 

\begin{table}[htb]
\begin{center}
\begin{tabular}{|c|c|c|} \hline
Architecture     & Original LeNet-5  & Modified LeNet-5 \\ \hline
1st Conv Layer Kernel Size & $5 \times 5 \times 1$   &  $5 \times 5 \times 3$      \\ \hline
1st Conv Layer Kernel No.  & $6$  &  $32$     \\ \hline
2nd Conv Layer Kernel Size & $5 \times 5 \times 6$   &  $5 \times 5 \times 32$      \\ \hline
2nd Conv Layer Kernel No.  & $16$  & $64$  \\ \hline
1st FC Layer Filter No.    & $120$ & $200$ \\ \hline
2nd FC Layer Filter No.    & $84$  & $100$ \\ \hline
Output Node No.            & $10$  & $10$  \\ \hline
\end{tabular}
\end{center}
\caption{Comparison of the original and the modified LeNet-5
architectures.}\label{table:mLeNet-5}
\end{table}

\begin{table}[htb]
\begin{center}
\begin{tabular}{|c|c|c|} \hline
Datasets  &     MNIST   & CIFAR-10  \\ \hline
FF        &    97.2\%  &  62\%  \\ \hline
Hybrid    &    98.4\%  &  64\%  \\ \hline
BP        &    99.9\%  &  68\%  \\ \hline
\end{tabular}
\end{center}
\caption{Comparison of testing accuracies of BP and FF designs
for the MNIST and the CIFAR-10 datasets.}\label{table:classification}
\end{table}

\begin{figure}[htb]
\centering
\begin{subfigure}[b]{0.4\linewidth}
\centering
\includegraphics[width=\linewidth]{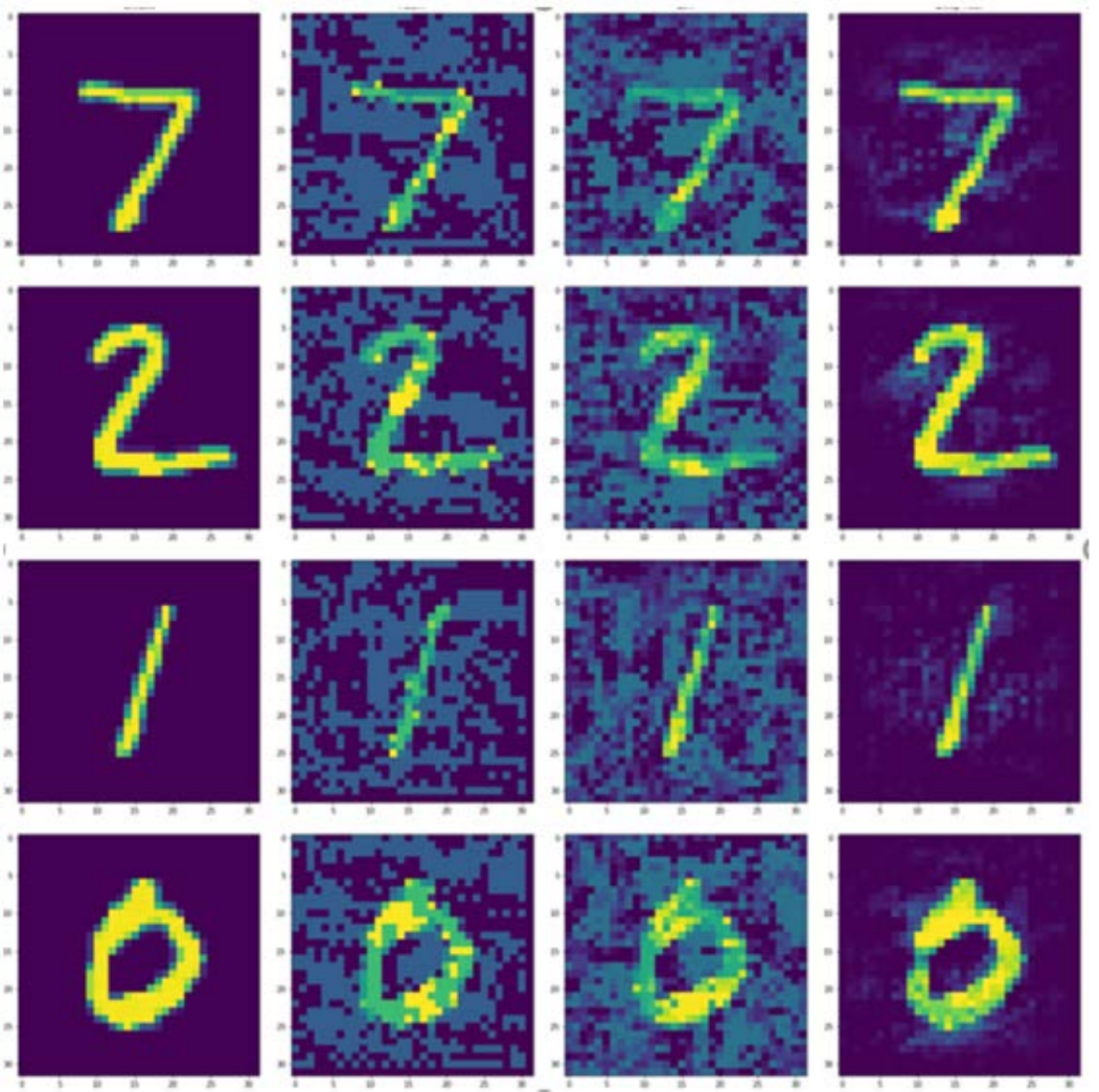}
\caption{MNIST}
\end{subfigure}
\begin{subfigure}[b]{0.4\linewidth}
\centering
\includegraphics[width=\linewidth]{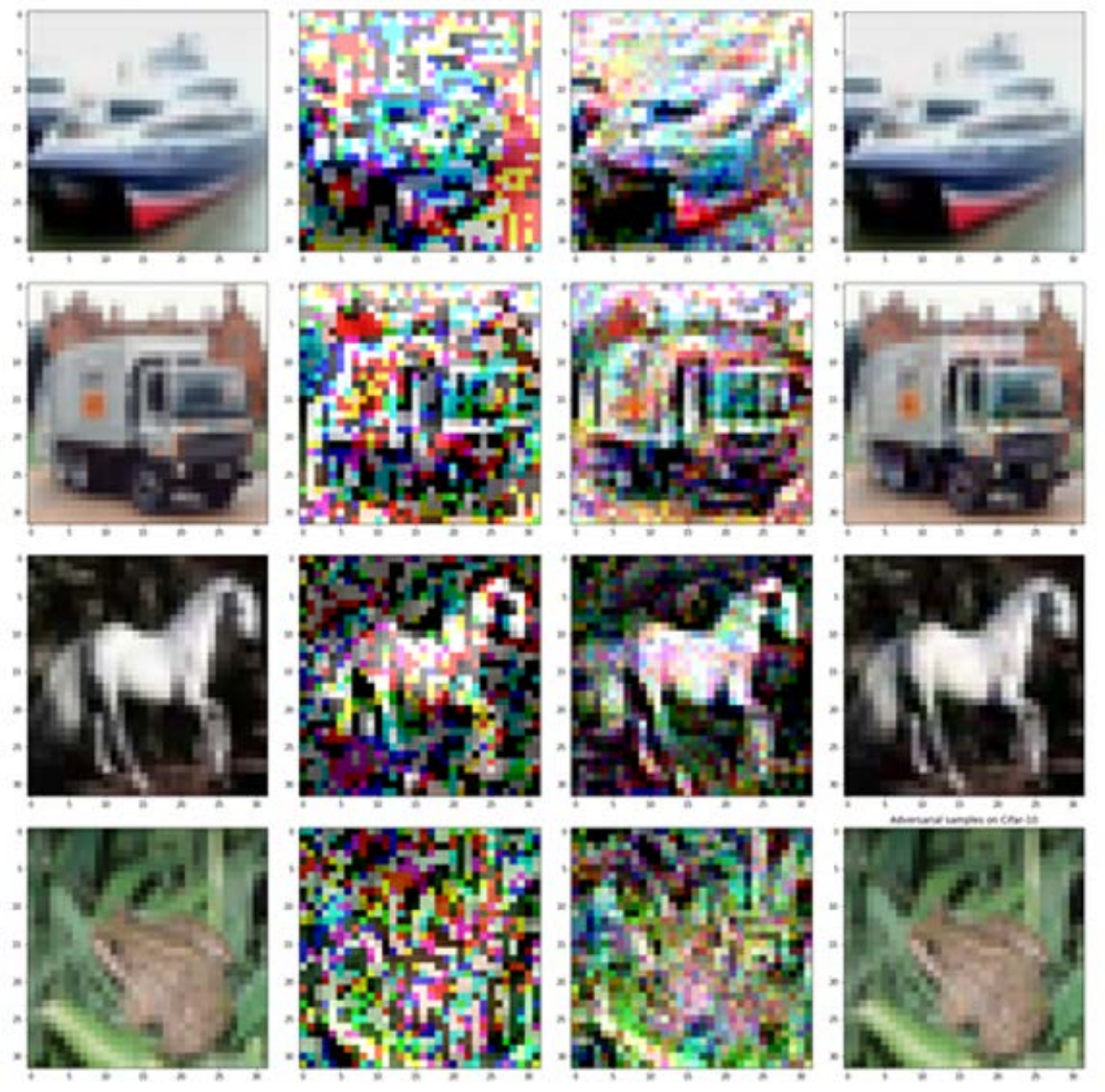}
\caption{CIFAR-10}
\end{subfigure}
\caption{Adversarial attacks to MNIST and CIFAR-10 datasets using
BP-designed model parameters (from left to right): original input
images, images attacked by the FGS, BIM and DeepFool.}\label{fig:attack}
\end{figure}

\begin{figure}[htb]
\centering
\begin{subfigure}[b]{0.4\linewidth}
\centering
\includegraphics[width=\linewidth]{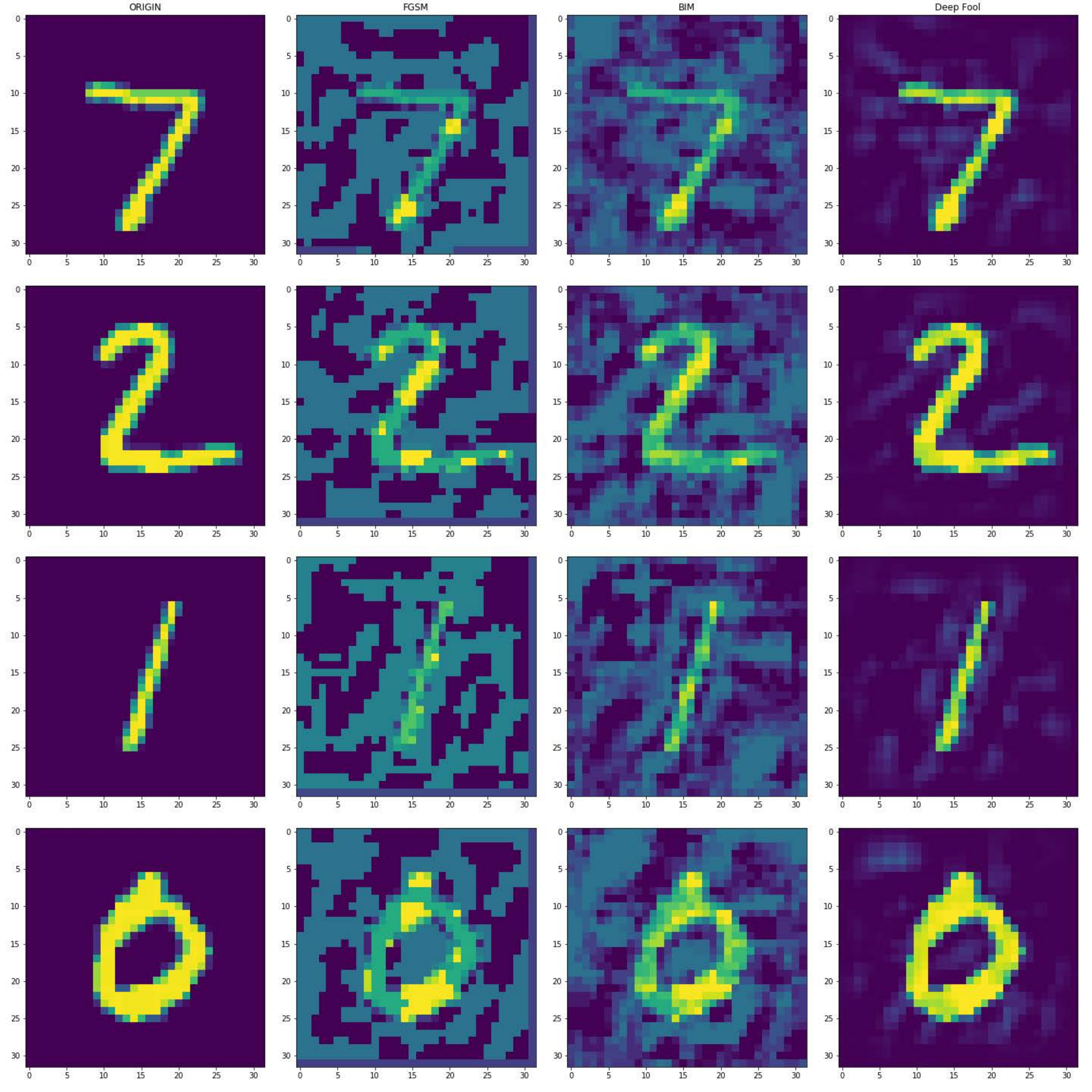}
\caption{MNIST}
\end{subfigure}
\begin{subfigure}[b]{0.4\linewidth}
\centering
\includegraphics[width=\linewidth]{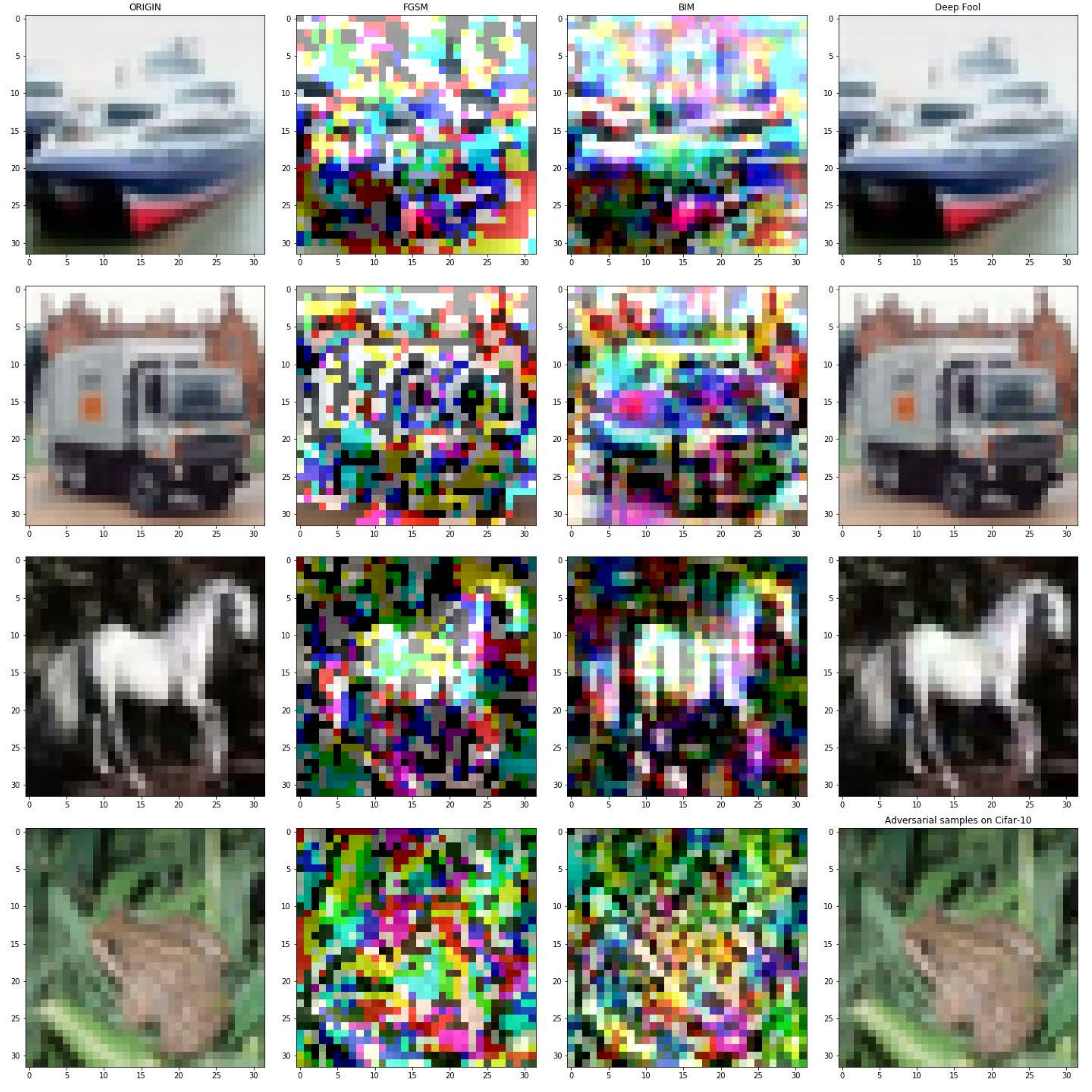}
\caption{CIFAR-10}
\end{subfigure}
\caption{Adversarial attacks to MNIST and CIFAR-10 datasets using
FF-designed model parameters (from left to right): original input
images, images attacked by the FGS, BIM and
DeepFool.}\label{fig:attack-2}
\end{figure}

{\bf Classification performance.} The testing accuracies of the BP and
the FF designs for the MNIST and the CIFAR-10 datasets are compared in
Table \ref{table:classification}. Besides, we introduce a hybrid design
that adoptes the FF design in the first two convolutional layers to
extract features. Then, it uses the multi-layer perceptron (MLP) the
last three FC layers with a built-in BP mechanism. 

By comparing the FF and the Hybrid designs, we observe the performance
degradation due to a poorer decision subnet without BP optimization.
We see a drop of 1.2\% and 2\% in classification accuracies for MNIST
and CIFAR-10, respectively. Next, we can see the performance degradation
primarily due to a poorer feature extraction subnet without BP
optimization. We see a drop of 1.5\% and 4\% for MNIST and CIFAR-10,
respectively.  Performance degradation due to the FF design is well
expected.  The performance gaps between the FF and the BP designs for
MNIST and CIFAR-10 are, respectively, 2.7\% and 6\%. 

\begin{table}[htb]
\begin{center}
\begin{tabular}{|c|c|c|c|} \hline
Attacks & CNN Design    &  MNIST     & CIFAR-10   \\ \hline
FGS     & BP            &  6 \%      &   15 \%    \\ \hline
FGS     & FF            &  56 \%     &   21 \%    \\ \hline \hline
BIM     & BP            &  1 \%      &   12 \%    \\ \hline
BIM     & FF            &  46 \%     &   31 \%    \\ \hline \hline
Deepfool& BP            &  2 \%      &   15 \%    \\ \hline
Deepfool& FF            &  96 \%     &   59 \%    \\ \hline 
\end{tabular}
\end{center}
\caption{Comparison of testing accuracies of BP and FF designs against
FGS, BIM and Deepfool three adversarial attacks targeting at the BP design.} 
\label{table:robustness}
\end{table}

\begin{table}[htb]
\begin{center}
\begin{tabular}{|c|c|c|c|} \hline
Attacks & CNN Design &  MNIST  & CIFAR-10   \\ \hline
FGS     & BP         &  34 \%  &   11 \%    \\ \hline
FGS     & FF         &  4 \%   &   6  \%    \\ \hline \hline
BIM     & BP         &  57 \%  &   14\%     \\ \hline
BIM     & FF         &  1 \%   &   12 \%    \\ \hline \hline
Deepfool& BP         &  97 \%  &   68 \%    \\ \hline
Deepfool& FF         &  2 \%   &   16 \%    \\ \hline 
\end{tabular}
\end{center}
\caption{Comparison of testing accuracies of BP and FF designs against
FGS, BIM and Deepfool three adversarial attacks targeting at the FF design.} 
\label{table:robustness-2}
\end{table}

{\bf Robustness against adversarial attacks.} Adversarial attacks have
been extensively examined since the pioneering study in
\cite{szegedy2013intriguing}. We examine three attacks: the fast
gradient sign (FGS) method \cite{goodfellow2014explaining}, the basic
iterative method (BIM) \cite{kurakin2016adversarial} and the Deepfool
method \cite{moosavi2016deepfool}.  

We generate adversarial attacks targeting at the BP and the FF designs
individually. Exemplary attacked images targeting at BP and FF designs
are shown in Fig. \ref{fig:attack} and Fig.  \ref{fig:attack-2},
respectively.  As shown in Figs. \ref{fig:attack} and
\ref{fig:attack-2}, the Deepfool provides attacks of least visual
distortion.  The quality of the FGS- and BIM-attacked images is poor.
Since one can develop algorithms to filter out poor quality images, we
should be more concerned with the Deepfool attack. 

The classification accuracies of BP and FF designs against attacked
images are compared in Tables \ref{table:robustness} and
\ref{table:robustness-2}.  The classification performance of BP and FF
designs degrades significantly against their individual attacks.  This
example indicates that, once CNN model parameters are known, one can
design powerful adversarial attacks to fool the recognition system yet
high-quality adversarial attacks have little effect on the HVS.  The
fact that CNNs are vulnerable to adversarial attacks of good visual
quality has little to do model parameters selection methodologies. It is
rooted in the end-to-end interconnection architecture of CNNs. To
mitigate catastrophic performance degradation caused by adversarial
attacks, one idea is to consider multiple models as explained below. 

{\bf Robustness enhancement via ensemble methods.} One can generate
different FF-designed CNN models using multiple initializations for the
k-means clustering in the multi-stage LSR models.  For example, we
construct the following three FF-designed CNNs that share the same
convolutional layer but different FC layers. 
\begin{itemize}
\item FF-1: Adopt the k-means++ initialization \cite{arthur2007k} in the
k-means clustering;
\item FF-2: Adopt the random initialization in the k-means clustering;
\item FF-3: Cluster five digits into 13 clusters and the other five
digits into 11 clusters in the design of the first AC layer (with 120
output dimensions) and adopt the k-means++ initialization.
\end{itemize}

\begin{table}[htb]
\begin{center}
\begin{tabular}{|c|c|c|c|} \hline
MNIST                   & FF-1       & FF-2           & FF-3                  \\ \hline
Deepfool attacking FF-1 & 2\%        &   79\%         &  81\%                 \\ \hline \hline
Deepfool attacking FF-2 & 78\%       &    2\%         &  81\%                 \\ \hline \hline
Deepfool attacking FF-3 & 79\%       &   80\%         &   2\%                 \\ \hline \hline
\end{tabular}
\end{center}
\caption{Comparison of MNIST testing accuracies of three FF designs,
against the Deepfool adversarial attacks targeting at FF-1 (the first
row), FF-2 (the second row) and FF-3 (the third row),
respectively.}\label{table:robustness-3}
\end{table}

\begin{table}[htb]
\begin{center}
\begin{tabular}{|c|c|c|c|} \hline
CIFAR-10                 & FF-1      & FF-2           & FF-3      \\ \hline
Deepfool attacking FF-1 & 16\%       &  47\%          & 48\%      \\ \hline 
Deepfool attacking FF-2 & 49\%       &  17\%          & 49\%      \\ \hline 
Deepfool attacking FF-3 & 48\%       &  49\%          & 15\%      \\ \hline \hline
\end{tabular}
\end{center}
\caption{Comparison of CIFAR-10 testing accuracies of three FF designs,
against the Deepfool adversarial attacks targeting at FF-1 (the first
row), FF-2 (the second row) and FF-3 (the third row), respectively.}
\label{table:robustness-4}
\end{table}

If the model parameters of the three FF designs are known, attackers can
design adversarial attacks by targeting at each of them individually.
The test accuracies of FF-1, FF-2 and FF-3 against the Deepfool attack
for the MNIST and the CIFAR-10 datasets are given in Tables
\ref{table:robustness-3} and \ref{table:robustness-4}, respectively. We
see that an adversarial attack is primarily effective against its target
design. To enhance robustness, we may adopt an ensemble method by fusing
their classification results (e.g., majority voting, bagging, etc.).
This is a rich topic that goes beyond the scope of this work. We will
report our further investigation in a separate paper.  Although the
ensemble method applies to both BP and FF designs, the cost of building
an ensemble system of multiple FF networks is significantly lower than
that of multiple BB networks. 

\section{Discussion}\label{sec:discussion}

Some follow-up discussion comments are provided in this section.  First,
comparisons between FF and BP designs are made in Sec.
\ref{subsec:comparison}. Then, further insights into the BP and FF
designs are given in Sec.  \ref{subsec:relationship}. 

\subsection{General comparison}\label{subsec:comparison}

\begin{table}[thb]
\begin{center}
\footnotesize
\begin{tabular}{|c|c|c|} \hline
Design          & BP                         & FF                       \\ \hline
Principle       & System optimization centric& Data statistics centric  \\ \hline
Math. Tools     & Non-convex optimization    & Linear algebra and statistics \\ \hline
Interpretability& Difficult                  & Easy                     \\ \hline
Modularity      & No                         & Yes                      \\ \hline
Robustness      & Low                        & Low                      \\ \hline
Ensemble Learning & Higher Complexity        & Lower Complexity         \\ \hline
Training Complexity  & Higher                & Lower                    \\ \hline
Architecture    & End-to-end network         & More flexible            \\ \hline
Generalizability& Lower                      & Higher                   \\ \hline
Performance     & State-of-the-art           & To be further explored   \\ \hline
\end{tabular}
\end{center}
\caption{Property comparison of BP and FF designs.}\label{table:comparison}
\end{table}

The properties of FF and BP designs are compared in various aspects in
Table \ref{table:comparison}. They are elaborated below. 

{\bf Principle.} The BP design is centered on three factors: a set of
training and testing data, a selected network architecture and a
selected cost function at the output end for optimization. Once all
three are decided, the BP optimization procedure is straightforward. The
datasets are determined by applications.  Most research contributions
come from novel network architectures and new cost functions that
achieve improved target performance.  In contrast, the FF design
exploits data statistics (i.e. the covariance matrix) to determine a
sequence of spatial-spectral transformations in convolutional layers
with two main purposes -- discriminant dimension generation and
dimension reduction. The PCA-based convolution and maximum spatial
pooling can be well explained accordingly. No data labels are needed in
the FF design of convolutional layers. After that, the FC layers provide
sequential LSR operations to enable a multi-stage decision process. Data
labels are needed in the FF design of FC layers.  They are used to form
clusters of data of the same class to build the LSR models. Clustering
is related to the sample distribution in a high-dimensional space. It is
proper to examine clustering and LSR from a statistical viewpoint. 

{\bf Mathematical Tools.} The BP design relies on the stochastic
gradient descent technique in optimizing a pre-defined cost function.
When a CNN architecture is deeper, the vanishing gradient problem tends
to occur. Several advanced network architectures such as the ResNets
\cite{He_2016_CVPR} were proposed to address this issue. The
mathematical tool used in this work is basically linear algebra.  We
expect statistics to play a significant role when our focus moves to
training data collection and labeling for a certain task. 

{\bf Interpretability.} As stated in Sec. \ref{sec:introduction},
interpretability of CNNs based on the BP design (or CNN/BP in short) was
examined by researchers, e.g., \cite{zhang2017interpretable,
wang2018interpret}.  Although these studies do shed light on some
properties of CNN/BP, a full explanation of CNN/BP is very challenging.
CNNs based on the FF design (or CNN/FF in short) are mathematically
transparent. Since the filter weights selection strategy in CNN/FF is
different from that in CNN/BP, our understanding of CNN/FF is not
entirely transferable to that of CNN/BP. However, we can still explain
the benefit of the multi-layer CNN architecture. Furthermore, the
connection between FF and BP designs will be built by studying
cross-entropy values of intermediate layers in Sec.
\ref{subsec:relationship}. 

{\bf Modularity.} The BP design relies on the input data and the output
labels. The model parameters at each layer are influenced by the
information at both ends. Intuitively speaking, the input data space has
stronger influence on parameters of shallower layers while the output
decision space has has stronger impact on parameters of deeper layers.
Shallower and deeper layers capture low-level image features and
semantic image information, respectively. The whole network design is
end-to-end tightly coupled. The FF design decouples the whole network
into two modules explicitly. The subnet formed by convolutional layers
is the data representation (or feature extraction) module that has
nothing to do with decision labels. The subnet formed by FC layers is
the decision module that builds a connection between the extracted
feature space and the decision label space. This decoupling strategy is
in alignment with the traditional pattern recognition paradigm that
decomposes a recognition system into the ``feature extraction" module
and the `classification/regression" module. 

{\bf Robustness.} We showed that both BP and FF designs are vulnerable
to adversarial attacks in Sec. \ref{sec:experiments} when the network
model is fixed and known to attackers. We provided examples to
illustrate that adversarial attacks lead to catastrophic performance
degradation for its target network but not others.  This suggests the
adoption of an ensemble method by fusing results obtained by multiple FF
networks.  The cost of generating multiple network models by changing
the initialization schemes for the k-means clustering in the FC layers
is very low. This is an interesting topic worth further investigation. 

{\bf Training Complexity.} The BP is an iterative optimization process.
The network processes all training samples once in an epoch.  Typically,
it demands tens or hundreds of epochs for the network to converge. The
FF design is significantly faster than the BP design based on our own
experience. We do not report the complexity number here since our FF
design is not optimized and the number could be misleading.  Instead, we
would like to say that it is possible to reduce complexity in the FF
design using statistics. That is, both PCA and LSR can be conducted on a
small set of training data (rather than all training data). Take the
filter design of the first convolutional layer as an example. The input
samples of the CIFAR-10 dataset are patches of size $5 \times 5 \times 3
=75$. We need to derive a covariance matrix of dimension $75 \times 75$
to conduct the PCA.  It is observed that the covariance matrix converges
quickly with hundreds of training images, which is much less than 50,000
training images in the CIFAR-10 dataset. One reason is that each
training image provides $28 \times 28=784$ training patches.
Furthermore, there exists an underlying correlation structure between
training images and patches in the dataset. Both high dimensional
covariance matrix estimation \cite{fan2008high} and regression analysis
with selected samples \cite{cameron2013regression} are topics in
statistics. Powerful tools developed therein can be leveraged to lower
the complexity of the FF design. 

{\bf Architecture.} The BP design has a constraint on the network
architecture; i.e. to enable end-to-end BP optimization.  Although we
apply the FF design to CNNs here, the FF design is generally applicable
without any architectural constraint. For example, we can replace the FC
layers with the random forest (RF) and the support vector machine (SVM)
classifiers. 

{\bf Generalizability.} There is a tight coupling between the data space
and the decision space in the BP design. As a result, even with the same
input data space, we need to design different networks for different
tasks. For example, if we want to conduct object segmentation,
classification and tracking as well as scene recognition and depth
estimation simultaneously for a set of video clips, we need to build
multiple CNNs based on the BP design (i.e., one network for one task)
since the cost function for each task is different. In contrast, all
tasks can share the same convolutional layers in the FF design since
they only depend on the input data space. After that, we can design
different FC layers for different tasks.  Furthermore, we can fuse the
information obtained by various techniques conveniently in the FF
design. For example, features obtained by traditional image processing
techniques such as edge and salient points (e.g. SIFT features) can be
easily integrated with FF-based CNN features without building larger
networks that combine smaller networks to serve different purposes. 

{\bf Performance.} The BP design offers state-of-the-art performances
for many datasets in a wide range of application domains. There might be
a perception that the performance of the BP design would be difficult to
beat since it adopts an end-to-end optimization process. However, this
would hold under one assumption. That is, the choice of the network
architecture and the cost function is extremely relevant to the desired
performance metric (e.g., the mean Average Precision in the object
detection problem). If this is not the case, the BP design will not
guarantee the optimal performance. The FF design is still in its infancy
and more work remains to be done in terms of target performance. One
direction for furthermore performance boosting is to exploit ensemble
learning. This is particularly suitable for the FF design since it is
easier to adopt multiple classifiers after the feature extraction stage
accomplished by multiple convolutional layers. 

\subsection{Further Insights}\label{subsec:relationship}

{\bf Degree of Freedom and Overfitting.} The number of CNN parameters is
often larger than that of training data. This could lead to overfitting.
The random dropout scheme \cite{JMLR:v15:srivastava14a} provides an
effective way to mitigate overfitting.  We should point out that the
number of CNN model parameters is not the same as its degree of freedom
in the FF design. This is because that the FF design is a sequential
process.  Filter weights are determined layer by layer sequentially
using PCA or LSR.  Since the output from the previous layer serves as
the input to the current layer, filter weights of deeper layers are
dependent on those of shallower layers. 

Furthermore, PCA filters are determined by the covariance matrix of the
input of the current layer. They are correlated.  The LeNet-5 has two
convolutional layers.  The inputs of the first and the second
convolutional layers have a dimension of $5 \times 5=25$ and $5 \times 5
\times 6=150$, respectively.  Thus, the covariance matrices are of
dimension $25 \times 25$ and $150 \times 150$, respectively. Once these
two covariance matrices are estimated, we can find the corresponding PCA
filters accordingly.  

As to FC layers, the dimension of the linear LSR matrix is roughly equal
to the product of the input vector dimension and the output vector
dimension. For example, for the first FC layer of the LeNet-5, the
number of parameters of its LSR model is equal to
\begin{equation}\label{eq:B}
(5 \times 5 \times 15 +1) \times 120=45,120, 
\end{equation}
where the DC channel is removed.  It is less than 60,000 training
images in the MNIST dataset.  The parameter numbers of the second FC
layer and the output layer are equal to $121 \times 84=10,164$ and $85
\times 10=850$, respectively. 

{\bf Discriminability of dimensions.} To understand the differences
between the BP and the FF designs, it is valuable to study the
discriminant power of dimensions of various layers.  The cross-entropy
function provides a tool to measure the discriminability of a random
variable by considering the probability distributions of two or multiple
object classes in this random variable.  The lower the cross entropy,
the higher its discriminant power. It is often adopted as a cost
function at the output of a classification system. The BP algorithm is
used to lower the cross entropy of the system so as to boost its
classification performance. We use the cross-entropy value to
measure the discriminant power of dimensions of an intermediate space. 

By definition, the cross-entropy function can be written as 
\begin{equation}\label{eq:cross-entropy}
L = \sum_{i=1}^{N_i} L_i,
\end{equation}
where $i$ is the sample index, $N_i$ is the total number of data
samples, and $L_i$ is the loss of the $i$th sample in form of
\begin{equation}\label{eq:cross-entropy-i}
L_i = - \sum_{c=1}^{N_c} y_{i,c} \log p_{i,c},
\end{equation}
where $c$ is the class index, $N_c$ is the total number of object
classes, $p_{i,c}$ is the probability for the $i$th sample in class $c$
and $y_{i,c}$ is a binary indicator ($0$ or $1$). We have $y_{i,c}=1$ if
it is a correct classification. Otherwise, $y_{i,c}=0$. In practice, we
only need to sum up all correct classifications in computing $L_i$. 

To compute the cross-entropy value in Eq. (\ref{eq:cross-entropy}), we
adopt the k-means clustering algorithm, partition samples in the target
dimension into $Q$ intervals, and use the majority voting rule to
predict its label in each interval. The majority voting rule is adopted
due to its simplicity.  Since we have the ground-truth of all training
samples, we know whether they are correctly classified. Consequently,
the correct classification probability, $p_{i,c}$, can be computed based
on the classification results in all intervals. 

\begin{figure}[htb]
\centering
\includegraphics[width=\linewidth]{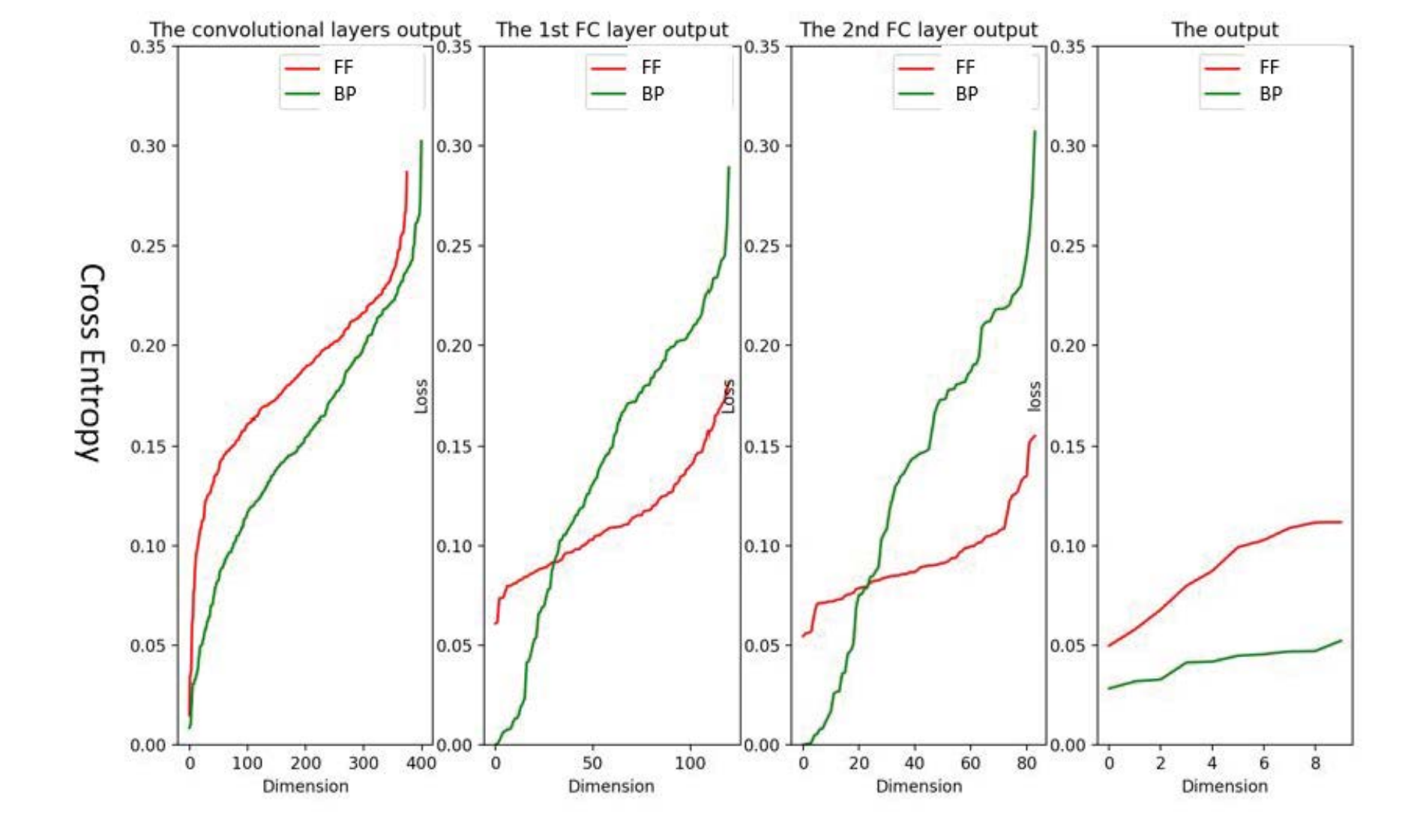}
\caption{Comparison of rank-ordered cross-entropy values of BP and FF
designs (from left to right): at the output of the second convolutional
layer, the first and the second FC layers and the output layer. This
illustrative example is obtained by applying the LeNet-5 to the MNIST
dataset.}\label{fig:centropy_comparison}
\end{figure}

Without loss of generality, we use the LeNet-5 applied to the MNIST
dataset as an example and plot rank-ordered cross-entropy values of four
vector spaces in Fig. \ref{fig:centropy_comparison}. They are: the
output of the second convolutional layer (of dimension $5 \times 5
\times 15 = 375$), the first and the second FC layers and the output
layer. The dimensions of the last three are 120, 84 and 10,
respectively. We omit the plot of cross-entropy values of the first
layers since each dimension corresponds to a local $5 \times 5$
receptive field and their discriminant power is quite weak. 

We see from Fig. \ref{fig:centropy_comparison} that the cross-entropy
values of the FF design are higher than those of the BP design at the
output of the 2nd convolutional layers as shown in the leftmost
subfigure. This is because the FF design does not use any label
information in the convolutional layer. Thus, the discriminability of
dimensions (or features) of the FF design is poorer.  The cross-entropy
values of the FF design in the middle two subfigures are much flatter
than those of the BP design. Its discriminant power is more evenly
distributed among different dimensions of the first and the second FC
layers. The distributions of cross-entropy values of the BP design are
steeper with a larger dynamic range. The BP design has some favored
dimensions, which are expected to play a significant role in inference
in the BP-designed network. The cross-entropy values continue to go
lower from the shallow to the deep layers, and they become the lowest at
the output layer.  We also see that the difference between the first and
the second FC layers is not significant.  Actually, the removal of the
second FC layer has only a small impact to the final classification
performance. 

{\bf Signal representation and processing.} We may use an analogy to
explain the cross-entropy plots of the BP algorithm.  For a given
terrain specified by a random initialization seed, the BP procedure
creates a narrow low-cross-entropy path that connects the input and the
output spaces through iterations. Such a path is critical to the
performance in the inference stage. 

\begin{table}[htb]
\begin{center}
\footnotesize
\begin{tabular}{|c|c|c|} \hline
Representation  & BP                        & FF               \\ \hline
Analogy         & Matched Filtering         & Projection onto Linear Space  \\ \hline
Search          & Iterative optimization    & PCA and LSR      \\ \hline
Sparsity        & Yes                       & No               \\ \hline
Redundancy      & Yes                       & No               \\ \hline
Orthogonality   & No                        & Yes              \\ \hline
Signal Correlation     & Stronger           & Weaker           \\ \hline
Data Flow       & Narrow-band               & Broad-band       \\ \hline
\end{tabular}
\end{center}
\caption{Comparison of BP and FF designs from the signal 
representation and processing viewpoint.}\label{table:sp}
\end{table}

We compare the BP and the FF designs from the signal representation and
processing viewpoint in Table \ref{table:sp}. The BP design uses an
iterative optimization procedure to find a set of matched filters for
objects based on the frequency of visual patterns and labels. When cat
faces appear very frequently in images labeled by the cat category, the
system will gradually able to extract cat face contours to represent the
cat class effectively. The FF design does not leverage labels in finding
object representations but conducts the PCA. As a result, it uses values
projected onto the subspace formed by principal components for signal
representation. The representation units are orthogonal to each other in
each layer.  The representation framework is neither sparse nor
redundant.  Being similar to the sparse representation, the BP design
has a built-in dictionary.  As compared to the traditional sparse
representation, it has a much richer set of atoms by leveraging the
cascade of multi-layer filters.  The representation framework is sparse
and redundant. Signal correlation is stronger in the BP design and
weaker in the FF design. Finally, the representation can be viewed as
narrow-band and broad-band signals in the BP and FF designs,
respectively. 

\begin{figure}[htb]
\centering
\includegraphics[width=0.45\linewidth]{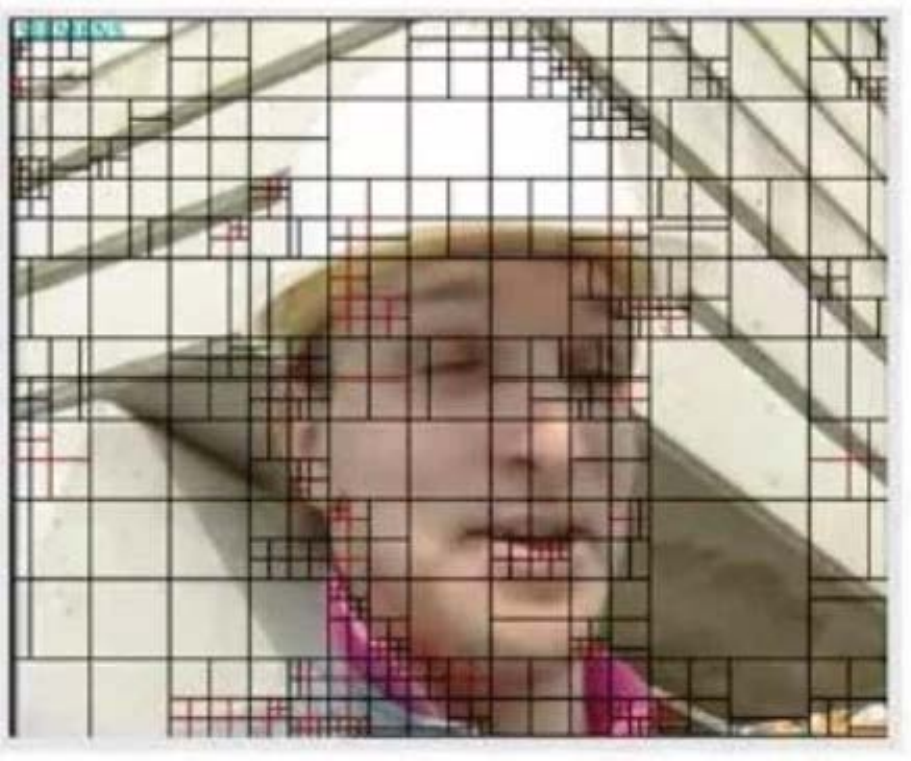}
\caption{Hierarchical image segmentation based on the HEVC 
video coding tool.}\label{fig:HEVC}
\end{figure}

{\bf Heterogeneous spatial-spectral filtering.} Although CNNs have
multiple filters at a convolutional layer, they apply to all spatial
locations. It is designed to capture the same pattern in different
locations. However, it is not effective for large images with
heterogeneous regions. As shown in Fig.  \ref{fig:HEVC}, an image can be
segmented into homogeneous regions with the HEVC coding tool
\cite{sullivan2012overview}.  Codebooks of PCA filters of different
sizes can be developed in the FF design.  PCA filters of the first
convolutional layers describe the pixel combination rules while those of
the second, third and higher convolutional layers characterize the
combination rules of input spatial-spectral cuboids.  The
segmentation-guided FF-CNN design is natural and interesting. 

\section{Conclusion and Future Work}\label{sec:conclusion}

An interpretable CNN design based on the FF methodology was proposed in
this work. It offers a complementary approach in CNN filter weights
selection. We conducted extensive comparison between these two design
methodologies. The new FF design sheds light on the traditional BP
design. 

As future extensions, it is worthwhile to develop an ensemble method to
improve classification performance and tackle with adversarial attacks
as discussed in Sec. \ref{sec:experiments}.  It will be interesting to
provide an interpretable design for advanced CNN architectures such as
ResNet, DenseNet and GANs.  Furthermore, it is beneficial to introduce
more statistical tools.  Today's CNN research has focused much on better
and better performances for benchmark datasets with more and more
complicated network architectures with respect to target applications.
Yet, it is equally important to examine the setup of a CNN solution from
the data viewpoint. For a given application or task, what data to
collect?  What data to label?  How many are sufficient?  Statistics is
expected to play a key role in answering these questions. 

The FF design methodology is still in its infancy.  Our work aims at
basic CNN research. It is of exploratory nature.  By providing a new
research and development topic with preliminary experimental results, we
hope that it will inspire more follow-up work along this direction. 

\section*{Appendix: Bias Selection}\label{subsec:saab}

We use ${\bf x}_{AC} \in R^N$ and ${\bf y}_{AC} \in R^K$ to denote input
and output flattened AC random vectors defined on 3D cuboids with two
spatial dimensions and one spectral dimension.  Anchor vectors, ${\bf
a}_k \in R^N$, $1 \le k \le K-1$, are the unit-length vectors obtained
by the principal component analysis of ${\bf x}_{AC}$ and used in the
Saab transform from ${\bf x}_{AC}$ to ${\bf y}_{AC}$.  We would like to
add a sufficiently large bias to guarantee that all response elements
are non-negative. 

The correlation output between ${\bf x}_{AC}$ and
anchor vector ${\bf a}_k$ can be written as
\begin{equation}\label{eq:corr1}
y_{k} = {\bf a}^T_k {\bf x}_{AC}, \quad k=1, \cdots, K-1.
\end{equation}
Clearly, ${\bf y}_{AC}=(0, y_1, \cdots, y_{K-1})^T$ is in the AC subspace
of $R^K$ since ${\bf x}$ is in the AC subspace of $R^N$.  Let ${\bf 1}= 
K^{-1/2} (1, \cdots , 1)^T$ be the constant-element unit vector in $R^K$.  
We add a constant displacement vector ${\bf d} = d {\bf 1}$ of length
$d$ to ${\bf y}_{AC}=(0, y_1, \cdots, y_{K-1})^T + {\bf d}$ and get a
shifted output vector ${\bf y}_d$ whose elements are
\begin{equation}\label{eq:corr2}
y_{d,k} = {\bf a}^T_k {\bf x} + d \sqrt{K}, \quad k=0, 1, \cdots, K-1.
\end{equation}
To ensure that $y_{d,k}$ is non-negative, we demand
\begin{equation}\label{eq:d1}
d \sqrt{K} \ge - {\bf a}^T_k {\bf x}, \quad k=0, 1, \cdots, K-1.
\end{equation}
The displacement length, $d$, can be easily bound by using the following inequality:
\begin{equation}\label{eq:ineq1}
- ||{\bf a}^T_k|| ||{\bf x}|| \le ||{\bf a}^T_k {\bf x}|| \le ||{\bf a}^T_k|| ||{\bf x}||.
\end{equation}
Since $||{\bf a}^T_k||=1$, we have
\begin{equation}\label{eq:ineq2}
- \max || {\bf x} || \le \max_{\bf x} || {\bf a}^T_k {\bf x} || \le \max || {\bf x} ||.
\end{equation}
By combining Eqs. (\ref{eq:d1}) and (\ref{eq:ineq2}), we obtain
\begin{equation}\label{eq:d2}
d \ge \frac{1}{\sqrt{K}} \max_{\bf x} || {\bf x} ||.
\end{equation}
Finally, we have the lower bound on $b_k$ as
\begin{equation}\label{eq:ab2}
b_k = d {\sqrt{K}} \ge \max_{\bf x} || {\bf x} ||, \quad k=0, 1, \cdots, K-1. 
\end{equation}
This result is repeated in Eq. (\ref{eq:d3}). Without loss of generality, we choose
\begin{equation}\label{eq:ab3}
b_k = \max_{\bf x} || {\bf x} || + \delta, \quad \delta > 0, \quad k=0, 1, \cdots, K-1,
\end{equation}
where $\delta$ is a small positive number in our experiment. 

\section*{Acknowledgment}

This material is partially based on research sponsored by DARPA and Air
Force Research Laboratory (AFRL) under agreement number
FA8750-16-2-0173. The U.S. Government is authorized to reproduce and
distribute reprints for Governmental purposes notwithstanding any
copyright notation thereon.  The views and conclusions contained herein
are those of the authors and should not be interpreted as necessarily
representing the official policies or endorsements, either expressed or
implied, of DARPA and Air Force Research Laboratory (AFRL) or the U.S.
Government. The authors would also like to give thanks to Dr. Pascal
Frossard and Dr. Mauro Barni for their valuable comments to the draft of
this work. 

\bibliographystyle{elsarticle-num} 
\bibliography{CNN2}

\end{document}